\documentclass[preprint,review,12pt]{elsarticle}

\usepackage{amssymb}
\usepackage{verbatim}
\usepackage{paralist,bbding,pifont}
\usepackage{subcaption}
\usepackage{array}
\usepackage{pifont}
\usepackage{soul, color, xcolor}
\usepackage{verbatim}
\usepackage{array}
\usepackage{makecell}
\usepackage{amsmath,amssymb,amsfonts}
\usepackage{graphicx}
\usepackage{textcomp}
\usepackage{xcolor}
\usepackage{booktabs}
\usepackage{bbding}
\usepackage{multirow}
\usepackage{colortbl}
\usepackage{amssymb}
\usepackage{graphicx}%
\usepackage{multirow}%
\usepackage{multicol}
\usepackage{amsmath,amssymb,amsfonts}%
\usepackage{amsmath}
\usepackage{amsthm}%
\usepackage{mathrsfs}%
\usepackage[title]{appendix}%
\usepackage{textcomp}%
\usepackage{manyfoot}%
\usepackage{booktabs}%
\usepackage{algorithm}%
\usepackage{algorithmicx}%
\usepackage{algpseudocode}%
\usepackage{listings}%
\usepackage{stfloats}
\usepackage{xcolor}
\usepackage{lscape}

\usepackage{booktabs}    
\usepackage{bbding} 
\usepackage[capitalize]{cleveref}

\definecolor{titlegrey}{rgb}{0.7, 0.7, 0.7}
\definecolor{contentgrey}{rgb}{0.8, 0.8, 0.8}
\begin{document}

\begin{frontmatter}



\title{$\mathrm{D^2}$Fusion: Dual-domain Fusion with Feature Superposition for Deepfake Detection}

\author[Durham]{Xueqi~Qiu}\ead{xueqi.qiu@durham.ac.uk}
\author[Durham]{Xingyu~Miao$^{\#}$}\ead{xingyu.miao@durham.ac.uk}
\author[Durham]{Fan~Wan}\ead{fan.wan@durham.ac.uk}
\author[Newcastle_Computing]{Haoran~Duan}\ead{haoranduan28@gmail.com}
\author[Newcastle_Computing]{Tejal~Shah}\ead{tejal.shah@newcastle.ac.uk}
\author[Newcastle_Computing]{Varun~Ojha\corref{mycorrespondingauthor}}\ead{varun.ojha@newcastle.ac.uk}
\author[Durham]{Yang Long\corref{mycorrespondingauthor}}\ead{yang.long@durham.ac.uk}
\author[Newcastle_Computing]{Rajiv~Ranjan}\ead{raj.ranjan@newcastle.ac.uk}
\address[Durham]{Department of Computer Science, Durham University, UK.}
\address[Newcastle_Computing]{School of Computing, Newcastle University, UK.}
\cortext[mycorrespondingauthor]{Corresponding author $^{\#}$ equal contribution}




\begin{abstract}
Deepfake detection is crucial for curbing the harm it causes to society. However, current Deepfake detection methods fail to thoroughly explore artifact information across different domains due to insufficient intrinsic interactions. These interactions refer to the fusion and coordination after feature extraction processes across different domains, which are crucial for recognizing complex forgery clues. Focusing on more generalized Deepfake detection, in this work, we introduce a novel bi-directional attention module to capture the local positional information of artifact clues from the spatial domain. This enables accurate artifact localization, thus addressing the coarse processing with artifact features. To further address the limitation that the proposed bi-directional attention module may not well capture global subtle forgery information in the artifact feature (e.g., textures or edges), we employ a fine-grained frequency attention module in the frequency domain. By doing so, we can obtain high-frequency information in the fine-grained features, which contains the global and subtle forgery information. Although these features from the diverse domains can be effectively and independently improved, fusing them directly does not effectively improve the detection performance. Therefore, we propose a feature superposition strategy that complements information from spatial and frequency domains. This strategy turns the feature components into the form of wave-like tokens, which are updated based on their phase, such that the distinctions between authentic and artifact features can be amplified. Our method demonstrates significant improvements over state-of-the-art (SOTA) methods on five public Deepfake datasets in capturing abnormalities across different manipulated operations and real-life. Specifically, in intra-dataset evaluations, $\mathrm{D^2}$Fusion surpasses the baseline accuracy by nearly 2.5\%. In cross-manipulation evaluations, it exceeds the baseline AUC by up to 6.15\%. In multi-source manipulation evaluations, it exceeds the SOTA methods by up to 14.62\% in P-value, 10.26\% in F1-score and 15.13\% in R-value. In cross-dataset experiments, it exceeds the baseline AUC by up to 6.25\%. For potential applications, $\mathrm{D^2}$Fusion can help improve content moderation on social media and aid forensic investigations by accurately identifying the tampered content.
\end{abstract}



\end{frontmatter}

\
\section{Introduction}\label{sec1}
Deepfake is an emerging facial video forgery technique which is used for creating fake videos based on AI technology. The videos generated by Deepfake can show people saying and doing things that never actually happened, and it is difficult for the human eye to distinguish the authenticity of these videos. Although Deepfake technology could be employed for productive endeavours such as film production, art style transformation, and virtual reality in the education field \cite{yadav2019deepfake}, Deepfake is often maliciously used for activities such as financial fraud, pornographic revenge, fake political news, and other activities that lead to property loss, damage the reputation of celebrities and misguide public opinion \cite{tolosana2020deepfakes}. Therefore, research around Deepfake detection is critical for cybersecurity, law, and politics, as well as for individual and social well-being.

Various Deepfake detection methods rely on a vanilla binary classifier to extract artifact features for detection \cite{chollet2017xception,he2016deep, tan2019efficientnet,li2020face,shiohara2022detecting}. This approach to coarse processing artifact features causes the network to predominantly focus on low-level manipulated trajectories \cite{yu2023fdml}, resulting in diminished precision in pinpointing the forged areas. Similarly, frequency information methods \cite{durall2019unmasking,frank2020leveraging,jeong2022frepgan} may ignore high-frequency signals, leading to the inability to capture some subtle forgery hints. In terms of combining different domain forgery traces, dual-branch network \cite{agarwal2021md}, dynamic graph \cite{wang2023dynamic}, and feature-disentangled representation learning \cite{miao2023f} are employed. However, the lack of sufficient generalization in processed features implies that while these methods may excel in detecting operation-specific artifacts, they often show a marked decline in performance when faced with unseen forgery techniques.

In this work, we present the $\mathrm{D^2}$Fusion framework, an innovative approach aimed at generalized facial forgery detection. To enhance artifact features, we introduce a bi-directional attention module. This module utilizes average pooling in both vertical and horizontal directions to extract local positional information from spatial domain, enabling precise localization of local artifacts. However, this attention module struggles to capture global details like texture and edge information. To address this, we integrate a fine-grained spectral attention module, employing discrete cosine transform (DCT) in multi-spectral partitioning. This approach retains high-frequency information including texture and edge information, thereby enhancing artifact feature details. To further strengthen the fusion of this complementary information from diverse domains, we propose a strategy of feature superposition. This strategy iteratively aggregates feature components according to their positional information. It amplifies the distinctiveness between the artifact feature and the authentic feature, thus making the features more generalized to various face manipulation algorithms and real-world scenarios. In summary, the contributions of this work are multi-fold:

\begin{compactitem}
\item {We investigate forged trajectories in artifact features from the feature-level perspective, and propose a bi-directional attention module that captures local information in features.}

\item{To obtain global detailed information in artifact features, we utilize multi-spectral components with DCT and propose a fine-grained frequency attention module.}

\item{To fuse information from different domains, we propose feature components in the form of wave-like tokens and update these tokens based on their phase, amplifying the difference between authentic and artifact features.}

\item{We show that our method outperforms state-of-the-art methods on five public datasets. Extensive experimental results demonstrate that our Deepfake detection model can capture abnormalities more effectively.}

\end{compactitem}

\section{Related Works}\label{sec2}

\subsection{Deepfake Generation}
Deepfake transformations can be categorized into two categories: face swapping and face reenactment. Face swapping can replace the face in the source image with the same face shape and features as the target face. Face reenactment is a face synthesis task in which the facial expressions and postures of the target face are transferred to the source face while preserving the appearance and details of the source face \cite{yu2021survey}.

In the early stage of Deepfake, based on computer graphics approaches \cite{lin2012face,nirkin2018face,smith2012joint}, the forgery methods exhibited significant drawbacks, like the loss of facial expressions and unnatural appearance \cite{rana2022deepfake}. After that, deep-learning based techniques are gaining popularity in generating synthetic media. The most common architecture of existing Deepfake generation \cite{li2019faceshifter,perov2020deepfacelab} is an auto-encoder with two separate decoders for facial identity exchange \cite{kingma2013auto}. Recent Deepfake research also takes advantage of Generative Adversarial Networks (GAN) \cite{choi2018stargan,karras2020analyzing,nirkin2019fsgan} for improved synthesis authenticity.

\begin{table}[t]
\caption{A comparison of recent research methods.}

\resizebox{\linewidth}{!}{\begin{tabular}{l|l|l|l|l|l|l}
\toprule
\rowcolor[HTML]{C0C0C0} 
{\color[HTML]{000000} Method}     & {\color[HTML]{000000} Year} & {\color[HTML]{000000} Category}   & {\color[HTML]{000000} \begin{tabular}[c]{@{}l@{}}Feature Type\end{tabular}} & \begin{tabular}[c]{@{}l@{}}Feature Pre-processing\end{tabular}                                                        & {\color[HTML]{000000} Strength}                                                                                                                & {\color[HTML]{000000} Weakness}                                                                                            \\ \midrule
Distorted Boundary\cite{li2018exposing}       & 2018               &                                                                               &                                                        &  \XSolidBrush                                                    & Utilize facial area distortion                                                                                                                        &                                                                \\ \cline{1-2} \cline{5-6} 
Face X-ray \cite{li2020face}    & 2020                     & \multirow{-2}{*}{\makecell[l]{Visual\\ artifacts} }                                           & \multirow{-2}{*}{Spatial}                                 &     \XSolidBrush                                                & Reveal the blending boundary                                                                                                                           & \multirow{-2}{*}{\makecell[l]{Low detection accuracy and limited\\ to basic visual artifacts   }}                                                                                       \\ \midrule

ID-Reveal \cite{cozzolino2021id}       & 2021              &                                                                               &                                                    &      \XSolidBrush                                                                                                           & \makecell[l]{Use prior biometric characteristics of \\a depicted identity }                                                                                           &                                                     \\  \cline{1-2} \cline{5-6} 
ICT \cite{dong2022protecting}   & 2022             &                                                                               &                                                 &       \XSolidBrush     & \makecell[l]{Combine transformer with identity\\ feature extraction }      &             \\ \cline{1-2} \cline{5-6} 
 \makecell[l]{Object\\ Representations \cite{bhaumik2023exploiting}  } & 2023        &           \multirow{-5}{*}{\makecell[l]{Object\\ detection} }                                & \multirow{-5}{*}{Spatial}                              &   \XSolidBrush                                                                                                                 & \makecell[l]{Look for object level coherence in\\ spatial dimensions}                                                                                                 &
\multirow{-5}{*}{\makecell[l]{Over-reliance on object features and\\ not as effective as deep learning based \\detection methods }   }                                                                   \\ \midrule  

MAT \cite{zhao2021multi}     & 2021                       &                                                                               &                                                   & \begin{tabular}[c]{@{}l@{}}Aggregate the different level\\ features with the attention maps\end{tabular} & \begin{tabular}[c]{@{}l@{}}Formulate deepfake detection as a\\ fine-grained classification problem\end{tabular}                                        &           \\ \cline{1-2} \cline{5-6} 
SBI \cite{shiohara2022detecting}    & 2022                        &                                              &    &     \XSolidBrush                                                                                                        &\makecell[l]{ Effective detection of most basic face \\swapping operations }                          &                                             \\ \cline{1-2} \cline{5-6} 

FADE \cite{tan2023deepfake}   & 2023                  &                                      &                             & Utilize multi-dependency graph                                                                                   & \makecell[l]{Can be integrated with some existing\\ frame-level methods}                          &             \\ \cline{1-2} \cline{5-6} 

CADDM\cite{dong2023implicit}   & 2023                  &                                                                               & \multirow{-5}{*}{Spatial}                               &\XSolidBrush                                                                                  & \makecell[l]{Address the implicit identity \\leakage issue}                                                                                                    &                                                                 \\ \cline{1-2} \cline{4-6} 
GFF \cite{luo2021generalizing}      & 2021                     &   &   & \makecell[l]{Residual guided spatial attention}                                                                   & \begin{tabular}[c]{@{}l@{}}Utilize
image noises\end{tabular}                                                        &  \\ \cline{1-2} \cline{5-6} 

FrepGAN \cite{jeong2022frepgan}    & 2022             &          &    &                                                                 \XSolidBrush                                                   & Not limited to the training settings                                                                                                                   &                                       \\ \cline{1-2} \cline{5-6} 
E-TAD \cite{gao2024texture} &2024 &  & &\XSolidBrush   &\makecell[l]{Focus on texture inconsistencies} &
\\ 
\cline{1-2} \cline{5-6} 

FreqNet \cite{tan2024frequency}   & 2024                 &                                                                               &                                                         & Frequency convolutional layer                                                                                     & \begin{tabular}[c]{@{}l@{}}The detector can consistently prioritize\\ and focus on high-frequency information\end{tabular}                             &                                                                                   \\ \cline{1-2} \cline{5-6}

IDM \cite{zhu2024deepfake}      & 2024                     &                                                                               & \multirow{-5}{*}{Frequency}                       & \makecell[l]{Feature recomposition and \\residual calculation }                                                                   & \begin{tabular}[c]{@{}l@{}}Amplify the illumination inconsistency\end{tabular}                                                        & \multirow{-13}{*}{\makecell[l]{Dependent on specific domain artifacts \\and vulnerability to adversarial attacks}} \\ \cline{1-2} \cline{4-7} 

SFDG \cite{wang2023dynamic}      & 2023              &                                                                               &                                                         & Dynamic graph learning                                                                                            & \begin{tabular}[c]{@{}l@{}}Discover the relationships with a\\ graph-based relation-reasoning approach\end{tabular}                      &              \\ \cline{1-2} \cline{5-6} 

FDML \cite{yu2023fdml}    & 2023   &                                                                               &                                                       & Feature-disentangling                                                                                          & \begin{tabular}[c]{@{}l@{}}Only forgery-relevant features are used \end{tabular}                      &                                          \\ \cline{1-2} \cline{5-6} 

TAN-GFD \cite{zhao2023tan}&2023 & & & Multi-level adaptive noise mining &  
\makecell[l]{Analyze and utilize texture and noise \\information} & 

\\ \cline{1-2} \cline{5-6} 
 LSDA \cite{yan2024transcending}&2024  & \multirow{-20}{*}{\makecell[l]{Deep \\learning}  }                                            & \multirow{-6}{*}{Spatio-frequency}   & Distill knowledge &  
\makecell[l]{Enhance generalization by simulating \\forgery feature variations} & 
\multirow{-6}{*}{\makecell[l]{Generalization ability still needs\\ further improve.}}
\\ 
 \bottomrule
 \end{tabular}}
\end{table}

\subsection{Deepfake Detection}
Early Deepfake detection relied on visual artifacts, the images generated by GANs or Autoencoders must be further distorted to align the original faces in the source video. Such a transformation leaves obvious visual artifact cues of distorted facial shape \cite{li2018exposing} and blending boundary \cite{li2020face}. Especially, \cite{li2020face} is foundational in exposing visual forgery traces. It achieves detection effectiveness by focusing on the presence of fusion boundaries, thereby reducing computational load. However, detections based on visual artifacts are limited to basic face-warping trajectories, they can not deal with more complex deepfake operations. 

Another kind of method focuses on object detection by identifying inconsistencies between the subject and the background in the spatial dimension. \cite{cozzolino2021id} introduces the ID-Reveal, which leverages metric learning and adversarial training strategies to learn the facial dynamic features of specific individuals. \cite{dong2022protecting} introduces the Identity Consistency Transformer (ICT) and addresses the utilization of high-level semantic information by detecting identity inconsistencies between the inner and outer facial regions. \cite{bhaumik2023exploiting} utilizes vision transformers to extract object representations and detect object-level spatial inconsistencies both intra-frame and inter-frame. Object detection based on inconsistencies between the background and the subject offers strong interpretability. Still, it may overfit detection toward fixed object features and be less effective than deep learning based detection methods.

Deep learning based detectors are currently the primary approaches \cite{zhao2021multi,ding2021repvgg,shiohara2022detecting,tan2023deepfake,dong2023implicit}, offering more complex feature extraction and classification compared to other detection approaches. In a single spatial domain, represented by MAT \cite{zhao2021multi}, which aggregates the different level features with the attention maps and first formulates Deepfake detection as a fine-grained classification problem. SBI \cite{shiohara2022detecting} adopts EfficientNetB4 \cite{tan2019efficientnet} as the classifier and offers a data augmentation strategy that uses self-blended images for training data. However, SBI typically generates full-face images, thus remaining confined to specific types of deepfake generation. Subsequent work \cite{dong2023implicit} improved upon SBI by employing multi-scale face swapping to reduce reliance on identity information. Meanwhile, theoretical underpinnings for identifying forged images using frequency domain are established in \cite{durall2019unmasking, frank2020leveraging}. The work in \cite{durall2019unmasking} demonstrates that the mean amplitude across different frequency bands in genuine images differs from that in fake images, identifying frequency domain information as a valuable detection indicator. Further, the study in \cite{frank2020leveraging} extensively analyzes the frequency spectrum across various GAN-based forgery techniques using Discrete Cosine Transform (DCT). Increasingly, subsequent works have leveraged frequency domain information for detection \cite{jeong2022frepgan,luo2021generalizing,gao2024texture,tan2024frequency,zhu2024deepfake}, among which \cite{gao2024texture} can focus on texture inconsistencies and \cite{tan2024frequency} can consistently prioritize and focus on high-frequency information. However, the simplistic processing of single-domain information limits the detection capabilities of models and vulnerability to adversarial attacks.


Furthermore, a part of the works employs both the spatial and frequency information \cite{wang2023dynamic,yu2023fdml,zhao2023tan,yan2024transcending}, but their generalization ability still needs further improvement. \cite{wang2023dynamic} introduces the use of dynamic graphs to explore high-order relations between spatial and frequency features, which requires a great number of parameters to store the relationship between spatial and frequency domains. TAN-GFD \cite{zhao2023tan} combines multi-scale texture difference features and regional noise inconsistency features, when performing cross-dataset detection, its performance declines. The model in \cite{yu2023fdml} tries to automatically separate forgery-relevant features in the spatial domain and frequency domain with a feature-disentangling strategy but requires a priori knowledge as a guide. \cite{yan2024transcending} introduces a Deepfake detection method based on Latent Space Data Augmentation (LSDA), which expands the forgery space by constructing and simulating variations of forgery features within the latent space but also increases the computational workload.

{In general, the above methods still lead to the model relying on the specific manipulated patterns and scenarios, {we prefer to have the model learn the more generalized differences between authentic and artifact features across different domains with their properties, thereby enabling the model to achieve excellent detectability in different forgery operations and scenarios.}

\section{Method}
\subsection{Architecture}
\begin{figure*}[ht]%
\centering
\includegraphics[width=1.0\textwidth]{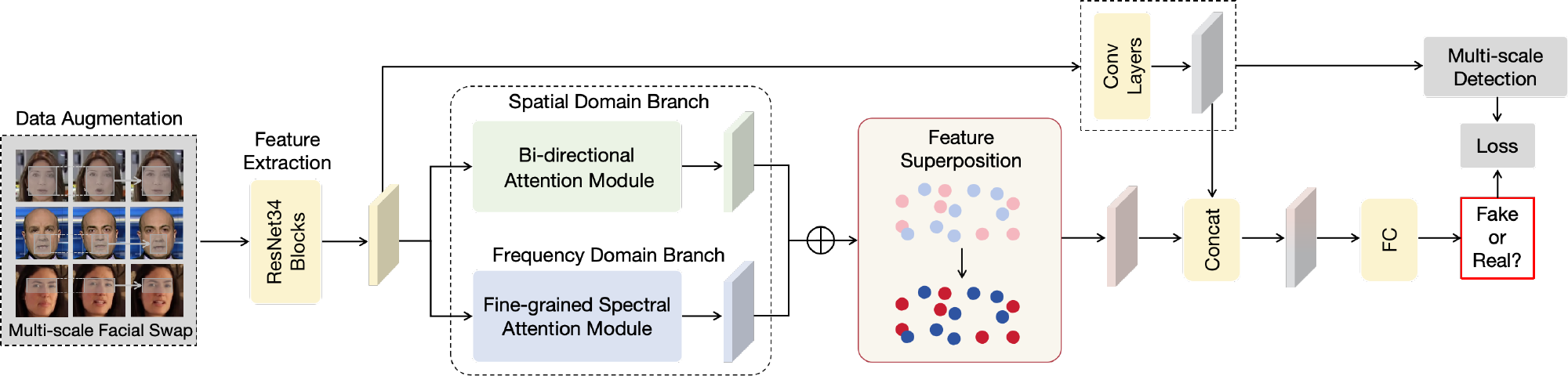}
\caption{The pipeline of our $\mathrm{D^2}$Fusion.}\label{pp}
\end{figure*}
In this section, we introduce the complete structure of $\mathrm{D^2}$Fusion (\Cref{pp}). Before facial forgery detection, we employ the multi-scale facial swap module \cite{dong2023implicit} for data augmentation. Subsequently, we utilize ResNet-34 as the backbone to extract facial features. 

The extracted features are then handled by the Dual-domain Attention area, comprising the bi-directional attention module and the fine-grained spectral attention module. The bi-directional attention module innovatively encodes features along both horizontal and vertical directions, subsequently generating an intermediate feature map through convolutional processing that integrates spatial information from both directions, effectively preserves the spatial relationships in forged images and accurately localizes manipulated areas. This is coupled with a fine-grained spectral attention module that converts the features into the frequency domain. It splits the feature into multi-spectral components with Discrete Cosine Transform (DCT), enhancing the focus on high-frequency details containing global artifact information.

Rather than simply learning from combined features, our network employs a feature superposition strategy for feature processing. Within this strategy, the positional information of tokens is defined as their phase, and their actual values correspond to the amplitude. This configuration facilitates the iterative fusion of tokens based on phase and amplitud, significantly boosting the discriminability of features. Additionally, we employ a multi-scale detection model \cite{dong2023implicit} as a shortcut to refine the detection outcomes.

\subsection{Preliminary}
We employ the multi-scale facial swap module \cite{dong2023implicit} to produce new forgery images for data augmentation. This module employs a randomly sized sliding window that aims to target the region most likely to contain artifacts:
\begin{equation}
    x_t,y_t=arg{\mathop{max}\limits_{x,y}}{\sum\limits_{i=x}^{x+h}} {\sum\limits_{j=y}^{y+w}}\mathrm{DSSIM}(I_f,I_s)_{i,j},
\label{dissim}
\end{equation}
where $x_{t}, y_{t}$ represents the top-left position of the sliding window on the image, $I_f$ is identified as the fake image, and $I_s$ as the source image. $\mathrm{DSSIM}(\cdot)$ \cite{wang2004image,loza2006structural} serves to quantify the differences between two images by evaluating brightness $l$, contrast $c$, and structure $s$:
\begin{equation}
\begin{split}    
\mathrm{DSSIM}(I_f,I_s)_{i,j}&=\frac{1}{1-[l(I_f,I_s)_{i,j}]^\alpha [c(I_f,I_s)_{i,j}]^\beta [s(I_f,I_s)_{i,j}]^\gamma},\\
l(I_f,I_s)_{i,j}&=(\frac{2\mu_{I_f}\mu_{I_s}+C_1}{\mu_{I_f}^2+\mu_{I_s}^2+C_1})_{i,j},\\
c(I_f,I_s)_{i,j}&=(\frac{2\phi_{I_f}\phi_{I_s}+C_2}{\phi_{I_f}^2+\phi_{I_s}^2+C_2})_{i,j}, \\
s(I_f,I_s)_{i,j}&=(\frac{\rho_{I_f,I_s}+C_3}{\rho_{I_f}\rho_{I_s}+C_{3}})_{i,j},
\end{split}
\end{equation}
where $\mu$ is mean, $\phi$ is standard deviation, $\rho$ is the covariant, and $C_1,C_2,C_3$ are all constants used to maintain $l,c,s$ stability. 

By omitting the region under the sliding window on the fake image, a mask $M$ is computed, facilitating the data augmentation with a new fake image $I_f^{'}$ with blending method $\xi$ as:
\begin{equation}
{I_f^{'}}=\xi (I_f,I_s,M).
\label{blending}
\end{equation}

\subsection{Dual-domain Attention}
\subsubsection{Bi-directional Attention Module}\label{co}

Previous work fails to consider the positional relationships between local facial parts, while we present the bi-directional attention (\Cref{block1}) that preserves spatial relationships in forged images and accurately localizes manipulated regions.

\begin{figure}[ht]%
\centering
\includegraphics[width=1\textwidth]{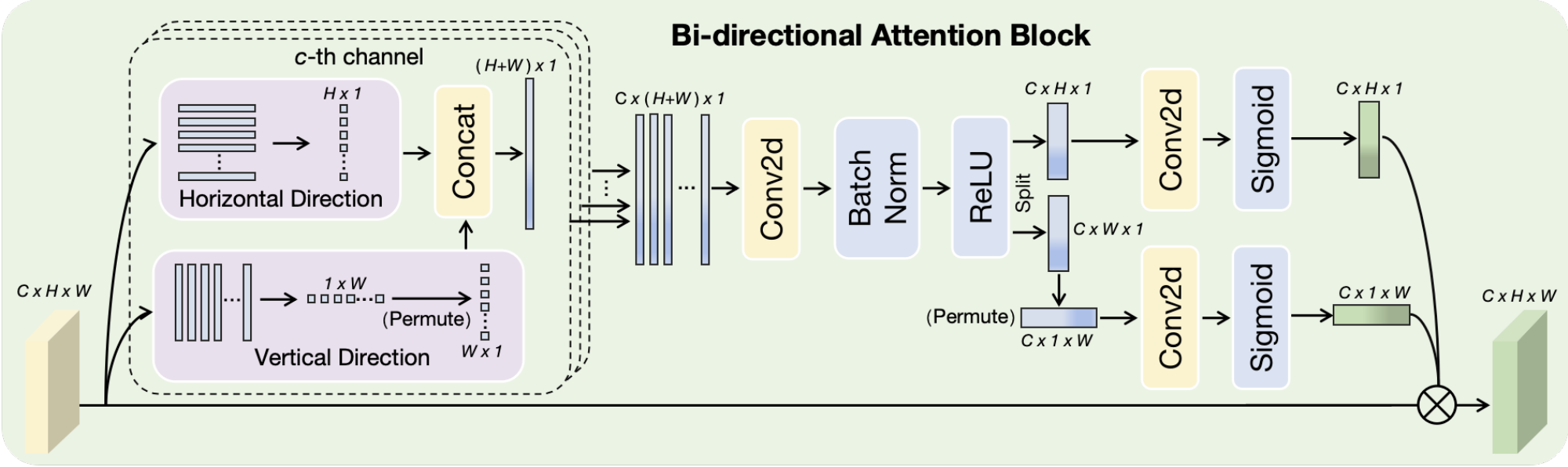}
\caption{The structure of bi-directional attention block.}\label{block1}
\end{figure}

Following \cite{hou2020strip}, given the input $X \in \mathbb{R}^{C\times{H}\times{W}}$, the module first encodes each channel along the horizontal and vertical directions, yielding a compact feature representation across different directions. For applying horizontal average pooling to $c-$th channel feature $x_c\in  \mathbb{R}^{{H}\times{W}}$, the module uses a spatial extent pooling kernel shaped $(1, W)$. The output $z_c^h$ of  $x_c$ at height $h$ can be expressed as:
\begin{equation}
    z_c^h=\frac{1}{W}{\sum\limits_{i=0}^{W-1}}{x_c^h},
    \label{hei}
\end{equation}
 and we obtain the transferred feature $Z^{H}_{c}=[z^{1}_{c},z^{2}_{c},\cdots,z^{H}_{c}], Z^{H}_{c}\in \mathbb{R}^{{H}\times{1}}$. 

Similarly, for applying vertical average pooling to $x_c$, the module uses another spatial extent pooling kernel shaped $(H, 1)$. The output of the $c-$th channel feature $x_c$ at width $w$ can be transformed into:
\begin{equation}
    y_c^w=\frac{1}{H}{\sum\limits_{i=0}^{H-1}}{x_c^w},
    \label{wid}
\end{equation}
and we obtain the transferred feature $Y^{W}_{c}=[y^{1}_{c},y^{2}_{c},\cdots,y^{W}_{c}], Y^{W}_{c}\in \mathbb{R}^{{1}\times{W}}$. We perform a permute operation on 
$Y^{W}_{c}$ to make $Y^{W}_{c}\in \mathbb{R}^{{W}\times{1}}$  for subsequent concatenation operation $\epsilon[\cdot,\cdot]$ :
\begin{equation}
    q_c=\epsilon[Z^{H}_{c},Y^{W}_{c}],
\end{equation}
$q_c \in \mathbb{R}^{{(H+W)}\times{1}}$ is the concatenation feature at $c$-th channel, and we can obtain the total concatenation feature $q_n=[q_1,q_2,\dots,q_C], q_n\in \mathbb{R}^{{C}\times{(H+W)}\times{1}}$. To enhance the expressive power of $q_n$, this module employ the $1 \times 1$ convolutional transformation function $F_1$ and ReLU function $\delta$, to obtain the intermediate feature map $f_{c}$:
\begin{equation}
    f_{c}=\delta(F_1(q_n)).
\end{equation}
The transformations described facilitate the intermediate feature map $f_{c}$ generation process, which encodes spatial information across both horizontal and vertical dimensions, ensuring stability and convergence.

To refocus each segment of the feature on a specific spatial direction, we split the feature map $f_{c}$ along the spatial dimension, yielding two separate tensors: $f^h_{c} \in \mathbb{R}^{C\times{H}\times{1}}$ and $f^w_{c} \in \mathbb{R}^{C\times{W}\times{1}}$. Also, for subsequent multiplication operations, we permute $f^w_{c} \in \mathbb{R}^{C\times{W}\times{1}}$ as $f^w_{c} \in \mathbb{R}^{C\times{1}\times{W}}$. Further expanding $f^h_{c}$ and $f^w_{c}$ using two more $1 \times 1$ convolutional transformations $F_h$ and $F_w$, with the Sigmoid function $\sigma$, and the reweighted output $X_{bi}$ from the bi-directional attention module is obtained as:
\begin{equation}
X_{bi}=X\times{\sigma(F_h(f^h_c))}\times{\sigma{(F_w(f^w_c))}}.
\end{equation}

\subsubsection{Fine-grained Spectral Attention Module}
\begin{figure}[t]%
\centering
\includegraphics[width=1\textwidth]{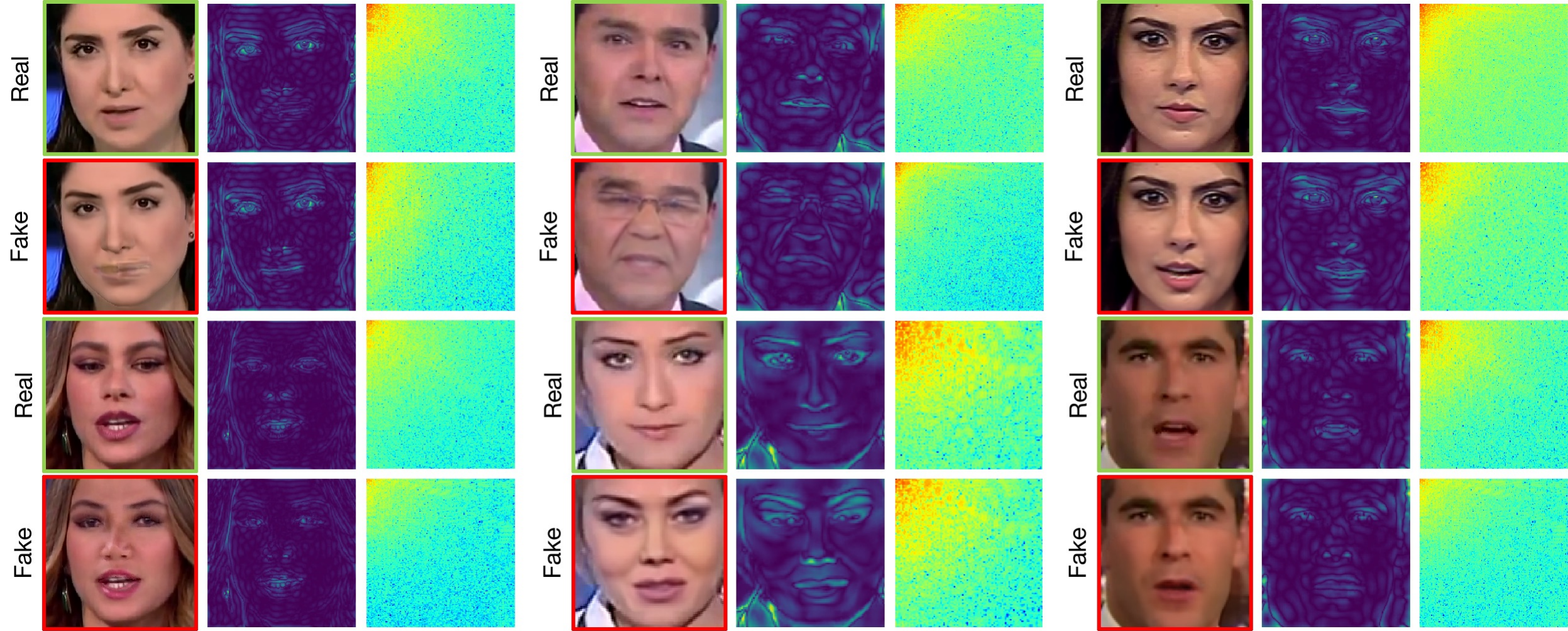}
\caption{This image displays the high-frequency information from real and fake images, with the corresponding spectrum. Generated images contain strong high frequencies components (visible as the more blue region in the lower right corner), while real images contain more lower frequency components (visible as the more brightened region in the top left corner.)}\label{high_f}
\end{figure}

The bi-directional attention module primarily identifies artifacts in the local facial components area. However, it ignores some additional global details, specifically those related to textures or edges, which are consistently found in high-frequency information, as displayed in \Cref{high_f}. To address this, we design a complementary attention module in the frequency domain (\Cref{block2}).

\begin{figure}[t]%
\centering
\includegraphics[width=1\textwidth]{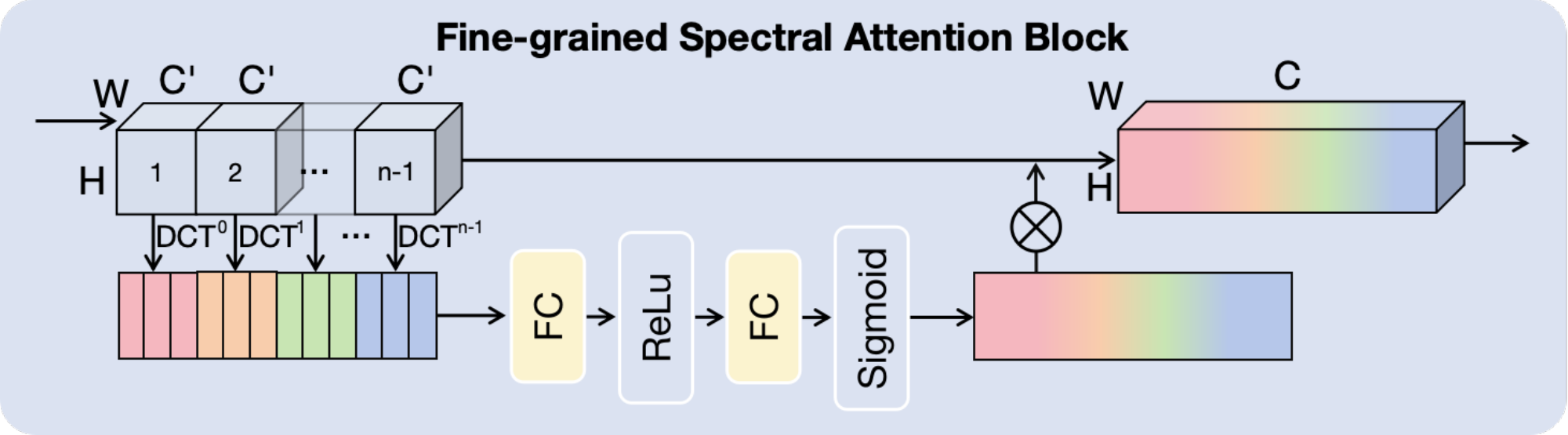}
\caption{The structure of fine-grained spectral attention block.}\label{block2}
\end{figure}

This module first splits the input feature $X$ into $n$ parts $\![X^0,X^1,\cdots,{X^{n-1}}]\!$ among channel dimension, in which $X^i \in \mathbb{R}^{C^{'}{\times{H}\times{W}}}(C^{'}=\frac{C}{n},i \in [0,1,\cdots,n-1])$. Thereafter, we utilizes the corresponding DCT \cite{ahmed1974discrete,rao2014discrete} to convert each input $X^i$ into the frequency domain for the reason that DCT exhibits strong energy compaction properties \cite{rao2014discrete}:
\begin{equation}
\label{m_dct}
\begin{split}
B^{u_{i},v_{i}}_{h,w}&=cos(\frac{\pi{h}}{H}(u_{i}+\frac{1}{2})cos(\frac{\pi{w}}{W}(v_{i}+\frac{1}{2}))\\
        {\nu}^i&=\sum_{h=0}^{H-1}\sum_{w=0}^{W-1}X^i_{:,h,w}B_{h,w}^{u_{i},v_{i}},
    \end{split}
\end{equation}
where $[u_{i}, v_{i}]$ are the frequency component indices corresponding to $X^i$, and ${\nu}^i \in \mathbb{R}^{C^{'}}$ is the $C^{'}$-th dimension vector after the compression. By doing so, the DCT is converged on a smaller scale, resulting in more reservation of high-frequency information. The whole compression spectral band $\kappa$ can be obtained after concatenation operation $\epsilon[\cdot,\cdots,\cdot]$:
\begin{equation}
\label{c_s}
    \begin{split}
    \kappa=\epsilon([{\nu}^{0},{\nu}^{1},\cdots,{\nu}^{n-1}]).
    \end{split}
\end{equation}

The weight $\kappa^{'}$ can be obtained with $\kappa$ after different fully connected functions and activation functions, and the final output $X_{sp}$ is:
\begin{equation}
X_{sp}=X\times{\kappa^{'}}.
\end{equation}

\subsection{Feature Superposition}
Although spatial features and frequency features can complement each other (e.g. the former contains the distribution of the local organs, the latter contains the global texture and edge information), the network still cannot effectively capture the feature differences between real and forgery. Therefore, we present the superposition strategy, which can enhance the fused features according to their positional information, thereby enabling the model to distinguish between various feature classes more effectively. Specifically, this strategy is divided into two stages: the implementation of a wave-like representation followed by phase-aware token-fusion (\Cref{block3}).

\begin{figure}[t]%
\centering
\includegraphics[width=1\textwidth]{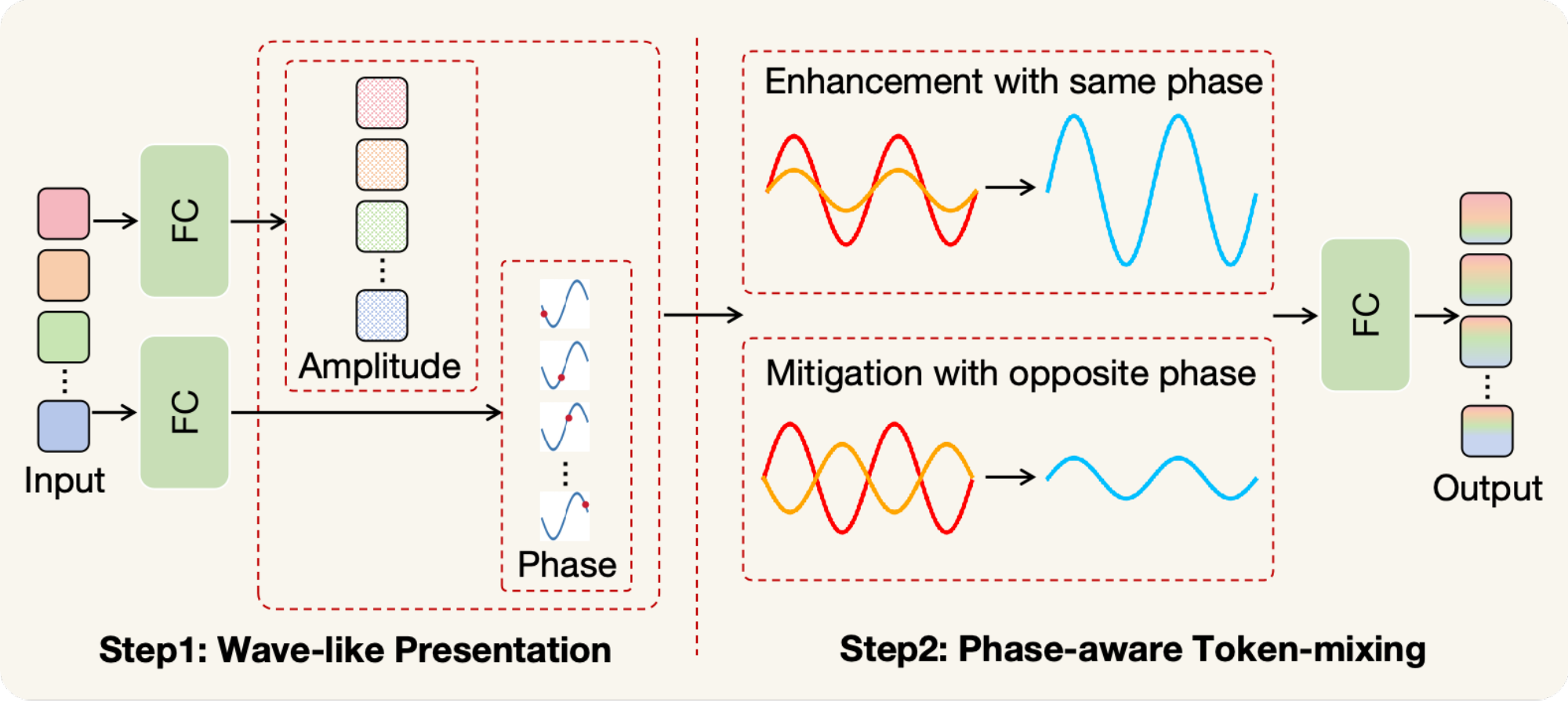}
\caption{Two-step processing in feature superposition.}\label{block3}
\end{figure}

\subsubsection{Wave-like Representation.} 

After the merging feature $X^{'} \in \mathbb{R}^{C\times{H}\times{W}}$ from $X_{bi} \in \mathbb{R}^{C\times{H}\times{W}}$ sum $X_{sp} \in \mathbb{R}^{C\times{H}\times{W}}$, we split $X^{'}$ into $m$ tokens $[x^{'}_{0}, x^{'}_{1}, \cdots, x^{'}_{m-1}]$ among height and width dimension, and we generate the amplitude and phase of each token $x^{'}_{i} \in \mathbb{R}^{C\times \frac{{H}\times{W}}{m}}$. In our wave-like term, we assign each token ``phase" and ``amplitude" attributes, which form the foundation for subsequent feature processing. This enables us to dynamically integrate tokens based on their semantic information, amplifying the distances between tokens with different attributes and clustering tokens with similar attributes to achieve improved classification outcomes.

\noindent\textbf{Amplitude:} The amplitude refers to the local information contained in the token and the assigned global facial information. In our work, we generate amplitude information through a plain channel-FC operation $\omega_1$ with the learnable parameters $W^{c}$, while the amplitude is a real-value feature, we employ an absolute value operation $\vert{\cdot}\vert$ to obtain the amplitude $\vert{z_{j}}\vert$ ($j \in [0,1,\cdots,m-1]$):
\begin{equation}\label{amp}
    \vert{z_{j}}\vert=\vert{{\omega_1}(x^{'}_{j},W^{c})}\vert.
\end{equation}

\noindent\textbf{Phase:} The phase $\theta_{j}$ is the current location of the token within a wave period. In our work, we utilize channel-FC to extract the positional information of this token in the face, thereby concretizing the phase. Specifically, we adopt another simple channel-FC $\omega_2$ with the learnable parameters $W^{q}$ as the phase estimation module: 
\begin{equation}\label{pha}
    \theta_{j}={\omega_2}(x^{'}_{j},W^{q}).
\end{equation}

\noindent\textbf{Wave-like Representation:} Each token is reinterpreted as a wave ${\tilde{z}}_j$, encapsulating both amplitude and phase information. For the purpose of embedding the wave-like token within a structure akin to that of the subsequent fully connected layers, the token undergoes expansion via the Euler formula and is expressed through its real and imaginary components:
\begin{equation} \label{expand}
    \tilde{z}_j=\vert z_j \vert \odot cos\theta_{j}+a\vert z_j \vert \odot sin\theta_j,
\end{equation}
where $a$ is the imaginary unit satisfying $a^2=-1$, $\vert z_j \vert \odot cos\theta_{j}$ is the real part, $a\vert z_j \vert \odot sin\theta_j$ is the imaginary part.

\subsubsection{Phase-aware Token-fusion}
Tokens can be categorized into two groups: authentic and artifact.  Current manipulated images are based on full-face region or local-face organs, without discrete generation of artifacts, so each category token has similar positional information formed as the phase.

When aggregating different tokens, supposing ${\tilde{z}_{r}}$ is the resultant wave of ${\tilde{z}_{k}}$ and ${\tilde{z}_{j}}$, its amplitude $\vert{{z}_{r}}\vert$ can be calculated as follows:
\begin{equation} \label{aggamp}
    \vert{z_{r}}\vert=\sqrt{{\vert{z_{k}}\vert}^{2}+{\vert{z_{j}}\vert}^{2}+2\vert{z_{k}}\vert \odot\vert{z_{j}}\vert\odot cos({\theta}_{j}-{\theta}_{i})}.
\end{equation}

The phase difference $(\theta_{j} - \theta_{i})$ between two tokens has a significant effect on the amplitude of the aggregated result $z_r$. When two tokens have the same phase ($\theta_j = \theta_k + m\pi, m \in [0, \pm2,\pm4, \cdots]$), they will be enhanced by each other, i.e., $\vert{z_r}\vert=\vert{z_k}\vert+\vert{z_j}\vert$. For the
opposite phase $(\theta_j = \theta_k + m\pi, m \in [\pm1,\pm3, \cdots])$, they will be weakened by each other, i.e., $\vert{z_r}\vert = \vert{\vert{z_k}\vert-\vert{z_j}\vert}\vert.$

The tokens interact with each other with their own amplitude and phase, and iterated token $\tilde{z}_r$ are then updated with the token-FC operation $\tau$:
\begin{equation} \label{mixing}
\begin{split}
     {\tilde{o}}_j&=\tau(\tilde{Z},W^{t})_{j}\\
    &=\sum_{r}{}W_{jr}^{t}\odot{\tilde{z_r}},
    \end{split}
\end{equation}
where ${\tilde{o}}_j$ is the updated tokens, $\tilde{Z} = [{\tilde{z}_1}, {\tilde{z}_2}, \cdots, {\tilde{z}_n}]$ denotes all the wave-like tokens in a layer, $W^{t}$
is the token-fusion weight. Following common quantum measurements \cite{braginsky1995quantum,jacobs2006straightforward}, we obtain the real-valued output $o_j$ by reweighting and summing the real and imaginary parts of ${\tilde{o}}_j$.
\begin{equation}\label{sum}
    o_{j}=\sum_{r}^{}(W_{jr}^{t}{z_{r}}\odot cos\theta_{r}+W_{jr}^{i}{z_{r}}\odot sin\theta_{r}).
\end{equation}

In the above equation, $W^{t}_{jr}, W^{i}_{jr}$ are learnable weights, the phase $\theta_r$ is dynamically adjusted according to the semantic content of the input data. To increase the representation capacity, the final output $P$ is obtained by transforming $o_{j}$ with another Channel-FC operation $\omega_3$ with weight $W^{p}$:
\begin{equation}\label{fin}
    P=\omega_{3}(o_{j}, W^{p}).
\end{equation}

\section{Experimental Results}\label{sec4}
In this section, we first introduce the implementation details and present our experimental result on the FaceForensics++ \cite{rossler2019faceforensics++}, Celeb-DF \cite{li2020celeb}, Celeb-DF-v2 \cite{li2020celeb}, the Deepfake Detection Challenge \cite{dolhansky2020deepfake}, and DeeperForensics-1.0 \cite{jiang2020deeperforensics}. Following this, we compare our proposed $\mathrm{D^2}$Fusion with other state-of-the-art methods, and an in-depth analysis is provided to understand our framework better.

\subsection{Datasets and Experimental Settings}
To evaluate the generalization ability of the proposed $\mathrm{D^2}$Fusion, we carry out the experiments on the following five public benchmark datasets: 
\\

\noindent\textbf{FaceForensics++(FF++):} The FF++ dataset is widely recognized as the most frequently used benchmark for detecting facial Deepfake videos. This dataset offers three compression rate options: original quality (RAW), high quality with light compression (HQ), and low quality with heavy compression (LQ). FF++ is composed of four sub-datasets, each corresponding to a specific generation method. For generating FF++ videos, two of these methods are based on computer graphics approaches: Face2Face (F2F) and FaceSwap (FS), while the other two rely on deep-learning approaches: DeepFakes (DF) utilizing auto-encoder and NeuralTextures (NT) using GANs. Within the FF++ dataset, there are 1000 real videos and each sub-dataset generates an additional 1000 fake videos, resulting in a total of 5000 videos.\\

\noindent\textbf{Celeb-DF (CD1):} The CD1 dataset comprises of both original and synthesized videos that exhibit visual quality similar to the videos commonly encountered online. The Celeb-DF dataset encompasses 408 original videos sourced from YouTube, featuring subjects of diverse ages, ethnic backgrounds, and genders. Additionally, within the Celeb-DF dataset, there are 795 Deepfake videos generated through synthesis based on these original videos.
\\

\noindent\textbf{Celeb-DF-v2 (CD2):} The CD2 dataset surpasses CD1 in terms of scale, containing 590 original videos along with 5639 corresponding Deepfake videos. Notably, the synthesized Deepfake videos within CD2 exhibit significant enhancements when compared to existing datasets. These improvements are particularly evident in areas such as stitching boundaries, color mismatches, inconsistent orientations, and other obvious visual artifacts \cite{nadimpalli2022improving}.
\\

\noindent \textbf{Deepfake Detection Challenge (DFDC):} The DFDC dataset represents a substantial dataset released for the Deepfake Detection Challenge competition. This dataset comprises of 19,154 authentic videos sourced from 3,426 compensated actors, alongside 100,000 counterfeit videos generated through a range of Deepfake techniques. The authentic videos within the DFDC dataset closely mirror real-life scenarios, while the areas with artifacts in its forgery videos exhibit greater precision compared to other datasets. 
\\

\noindent\textbf{DeeperForensics-1.0 (DFR):} The DFR dataset is a high-quality dataset specifically crafted for real-world face forgery detection. It encompasses 60,000 videos, with a staggering 17.6 million frames, in this dataset, there are 48,475 source videos and 11,000 manipulated videos. The dataset places a strong emphasis on realism, scale, and diversity, capturing real-world variations such as sub-optimal illumination, extensive occlusion of the target faces, and extreme head poses. Concretely, in DFR, the authors proposed a many-to-many end-to-end face-swapping technique known as the Deepfake Variational Auto-Encoder \cite{kingma2013auto} (DF-VAE) to generate Deepfake videos.
\\

\begin{table}[]
\caption{Publicly available datasets.}
\resizebox{\linewidth}{!}{\begin{tabular}{l|l|l|l|l|l}
\toprule
\rowcolor[HTML]{C0C0C0} 
{\color[HTML]{000000}Dataset  }               &{\color[HTML]{000000}FF++}   & {\color[HTML]{000000}CD1}                               &{\color[HTML]{000000}CD2}   &{\color[HTML]{000000}DFDC    }                & {\color[HTML]{000000}DFR    }                                                                                   \\ \midrule

Real/Fake sample size        & \begin{tabular}[c]{@{}l@{}}1000 real videos, \\ 4000 fake videos\end{tabular} & \begin{tabular}[c]{@{}l@{}}408 real videos, \\ 795 fake videos\end{tabular} & \begin{tabular}[c]{@{}l@{}}590 real videos, \\ 5639 fake videos\end{tabular} & \begin{tabular}[c]{@{}l@{}}19154 real videos, \\ 100000 fake videos\end{tabular} & \begin{tabular}[c]{@{}l@{}}50000 real videos, \\ 10000 fake videos\end{tabular} \\  \midrule

Sample catagory         & \begin{tabular}[c]{@{}l@{}}Face swapping\\ Face reenactment\end{tabular}        & Face swapping                    & Face swapping                                                                 & Face swapping                                                                      & Face reenactment                                                                  \\  \midrule

Size before preprocessing & \begin{tabular}[c]{@{}l@{}}640$\times$480, 1280$\times$720,\\ 1920$\times$1080\end{tabular}         & Various                          & Various                                                                       & 320$\times$240 - 3840$\times$2160                                                              & 1920$\times$1080                                                                        \\ \midrule
Size after preprocessing  & \multicolumn{5}{c}{224$\times$224}                                                                                                                                 \\  \midrule
Scenarios               & YouTube                                                                         & YouTube                          & YouTube                                                                       & Actors                                                                              & Actors                                                                             \\ \bottomrule                     
\end{tabular}}

\end{table}
\noindent\textbf{Implementation Details:} We train $\mathrm{D^2}$Fusion on a single NVIDIA RTX3090 GPU and use PyTorch to build it. We adopt ResNet34 \cite{he2016deep} as our backbone, which is pre-trained on the ImageNet dataset \cite{deng2009imagenet}. For each video, we evenly select 32 frames for testing and training. The Dlib is employed for face detection and extraction in the frames, and the cropped faces are resized to 224×224. Multi-scale facial swap sliding window scale is randomly selected from [40, 80], [80, 120], [120, 160], [224, 224]. 

During the training phase, we set the batch size to 64 and record results at epoch 100, 200, 300, 400, 500 to find the optimal checkpoint. We use Adam \cite{kingma2014adam} as our optimizer and the learning rate is set to $3.6\times{10^{-4}}$ at initialization, and decrease to $2\times{10^{-4}}$ at epoch 20, $1\times{10^{-4}}$ at epoch 40, $5\times{10^{-5}}$ at epoch 60, and $1\times{10^{-5}}$ at epoch 80 for fine-tuning. It is worth noting that in our experiments, all evaluation measures are computed at the frame level. 
\\

\noindent\textbf{Evaluation Metrics: }In our different experiments, we utilize a total of five types five evaluation metric: the area under the curve (AUC), accuracy (ACC), precision($P$), recall($R$), and F1 score($F1$):
\begin{equation}
    \begin{split}   
    AUC&=p(p(\frac{TP}{TP+FN})>p(\frac{FP}{TN+FP})),\\
    ACC&=\frac{TP+TN}{TP+TN+FP+FN}, \\
    P&=\frac{TP}{TP+FP},\\
    R&=\frac{TP}{TP+FN},\\
    F1&=2\frac{{P}\times{R}}{P+R},\\
    \end{split}
\end{equation}
$p(\cdot)$ denotes probability, $TP$, $TN$, $FP$ and $FN$ denote the counts of true positive, true negative, false positive, and false negative samples, respectively.

\subsection{Comparison with State-of-the-art Methods} \label{comparison}
We use various datasets to compare our method with other face forgery detection techniques to show its effectiveness and generalization performance. To make the comparison fair and complete, we reproduce some corresponding experiments using state-of-the-art (SOTA) methods under the same experiment setting. Note that in all results tables, we underline the second-best results while the best results of all listed methods are in bold. 
\begin{figure}[t]%
\centering
\includegraphics[width=1\textwidth]{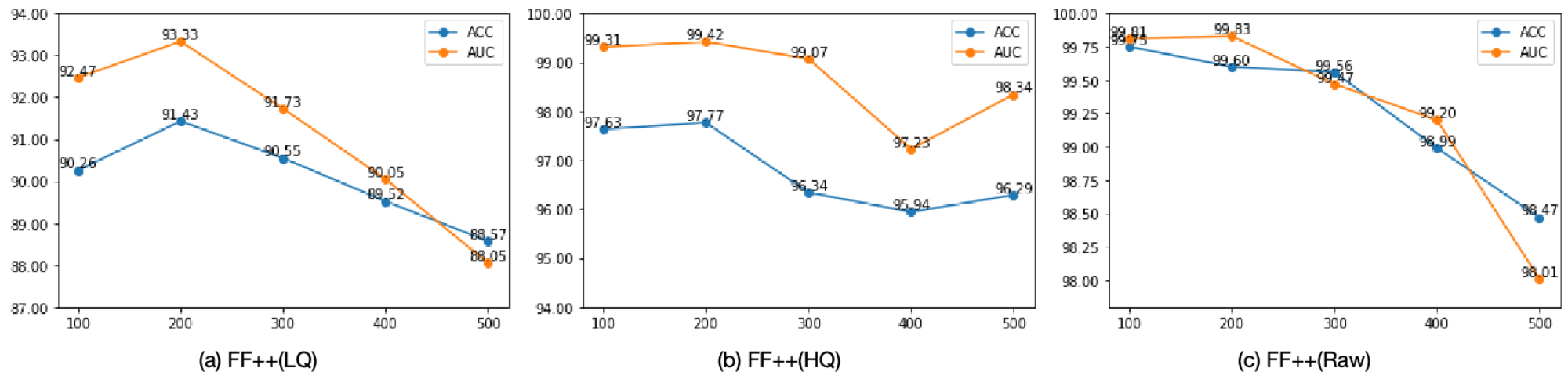}
\caption{Incremental comparison experiments using three compression rate versions of FF++ at different epochs.}\label{epoch}
\end{figure}

\begin{table}[t]
\centering
\caption{Intra-dataset evaluation with other methods on FF++ with different compression rate in terms of ACC (\%) and AUC (\%).}

\resizebox{1.0\linewidth}{!}{

\begin{tabular}{cccccccc}
\toprule
\multirow{2}{*}{Method}  & \multirow{2}{*}{Year}              & \multicolumn{2}{c}{LQ} & \multicolumn{2}{c}{HQ} & \multicolumn{2}{c}{RAW} \\ \cmidrule(lr){3-8} 
  &               & ACC          & AUC          &  ACC          & AUC           &  ACC          & AUC           \\ \midrule

Face X-ray  \cite{li2020face} &2020        &    -          & 61.60         &    -          & 87.35         &       -       &     -         \\

GFF \cite{luo2021generalizing} &2021           & 86.89        & 88.27        & 96.87        & 98.95        &       -       &   -           \\

MAT(EN-B4) \cite{zhao2021multi}&2021        &     88.69         &     90.40         &    \underline{97.60}          &  99.29           &    -          & -             \\

TAN-GFD \cite{zhao2023tan}&2023 & 87.42&89.26& 97.17& 99.21&-&-\\

CADDM  \cite{dong2023implicit} &2023       &      89.04        &  92.32       &      97.59            &  99.24            &     \underline{ 99.51}        &  \underline{ 99.78 }          \\

FADE(ResNet) \cite{tan2023deepfake}&2023& \underline{90.56} & \underline{93.37} & 97.55& 99.31 & -&  -\\  
IDM \cite{zhu2024deepfake} & 2024&- &\textbf{93.57} & - &\textbf{99.74}&-&-\\
\textbf{Ours} &2024             &   \textbf{   91.43 }       &      93.33        &   \textbf{97.77}           &  \underline{99.42}           &      \textbf{99.60}       &  \textbf{99.83}     \\ \bottomrule      

\end{tabular}
}
\label{tab1}

\end{table}

\begin{table}[ht]
\caption{Cross-manipulation evaluation with other methods on FF++(HQ) in terms of AUC (\%). In the \textit{Bias} column, we show the AUC change when compared to CADDM.}
\resizebox{1\linewidth}{!}{
\begin{tabular}{ccccccccc}\toprule
Train                                                                              & Test & \makecell[c]{EN-B4 \cite{tan2019efficientnet} \\(2019)} & \makecell[c]{MAT \cite{zhao2021multi}\\(2021)} & \makecell[c]{GFF  \cite{luo2021generalizing}\\(2021)} & \makecell[c]{DCL  \cite{sun2022dual}\\(2022)} & \makecell[c]{CADDM  \cite{dong2023implicit}\\(2023)} & \textbf{Ours} & Bias \\ \midrule
\multirow{4}{*}{\begin{tabular}[c]{@{}c@{}}F2F\\ (Computer Graphics)\end{tabular}} & F2F  & 99.20                                             & 99.13          & 99.10                  & 99.21                                    & \underline{99.67}          & \textbf{99.86}       & +0.19                   \\
         & FS   & 58.14   & 60.14                 & 61.30           & 59.58      & \textbf{64.07}                   & \underline{62.47}  & -1.60 \\
         & DF   & 84.52       & 86.15       & 89.23      & \textbf{91.91}           & 88.19         & \underline{89.50} & +1.31  \\
        & NT   & 63.71        & 64.59         & 64.77        & 66.67    & \underline{72.74}      & \textbf{75.23}  & +2.49     \\ \midrule
\multirow{4}{*}{\begin{tabular}[c]{@{}c@{}}FS\\ (Computer Graphics)\end{tabular}}  & F2F  & 67.69                                             & 66.39                                     & 68.72                                            & \underline{69.95}          & \textbf{73.28}          & 69.76       & -3.52    \\
             & FS   & 99.89          & 99.67       & 99.85         & 99.90           & \underline{99.91}            & \textbf{99.92} & +0.01       \\
  & DF   & 69.25                   & 64.13                                     & 70.21                                            & 74.80                                    & \underline{75.66}                         & \textbf{77.50}        & +1.84                  \\
& NT   & 48.61                                             & 50.10                                     & 49.91                                            & 52.60                                    & \textbf{59.94}                                           & \underline{58.45} & -1.49 \\ \midrule
\multirow{4}{*}{\begin{tabular}[c]{@{}c@{}}DF\\ (Auto-encoder)\end{tabular}}    & F2F  & 76.32                                             & 75.23          & 76.89                     & 77.13       & \underline{77.85}       & \textbf{77.88}            & +0.03 \\
   & FS   & 46.24         & 40.51               & 47.21                                            & 61.01                                    & \textbf{64.93}                                           & \underline{62.25}   & -2.68  \\
            & DF   & \underline{99.97}             & 99.92                 & 99.87                                            & \textbf{99.98}           & 99.94            & \textbf{99.98}  &+0.04 \\
    & NT   & 72.72        & 71.08    & 72.88  & \underline{75.01}     & 71.86                                  & \textbf{75.73} &+3.87\\ \midrule
\multirow{4}{*}{\begin{tabular}[c]{@{}c@{}}NT\\ (GAN)\end{tabular}}                & F2F  & 48.86                                             & 48.22                                     & 49.81                                            & 52.13                 & \underline{67.03}                      & \textbf{71.08}   &+4.05  \\
    & FS   & 73.05                                             & 75.33                                     & 74.31                                            & \underline{79.31}                                    & 74.60    & \textbf{80.75}   & +6.15                     \\
             & DF   & 85.99                                             & 87.23                                     & 88.49                     & 91.23                                    & \underline{92.40}  & \textbf{94.44}   &+2.04 \\
            & NT   & 98.25                                             & 98.66                                     & 98.77                  & 98.98                                    & \underline{99.09}            & \textbf{99.43}  & +0.34          \\ \bottomrule             
\end{tabular}}
\label{tab2}
\end{table}
\subsubsection{Intra-dataset Evaluation on FF++} \label{Intra-dataset evaluation on FF++}

We initially assess D$^2$Fusion at various epochs to determine the optimal checkpoint. In \Cref{epoch}, we conduct an incremental comparison and find that at epoch 200, the model achieves peak performance within the dataset across different compression rates. As the number of epochs increases, both ACC and AUC metrics show a decline.

We evaluate our method in comparison to previous detection methods using three versions of FF++. The comparative results are presented in Table \ref{tab1}. Overall, irrespective of the compression ratio applied to the video dataset, our proposed method consistently outperforms earlier visual artifact-based backbone networks such as Face X-ray, which can only detect the most basic face forgery operations.

In three versions of FF++, compared to the methods that use single domain information, in the case of raw quality images, Our method demonstrates a significant performance improvement, achieving a 0.09 increase in ACC and a 0.05 enhancement in AUC compared to CADDM \cite{dong2023implicit}. Similarly, MAT \cite{zhao2021multi} exhibits this deficiency with AUC decreased by 2.93. The superior performance of our approach contributes to its incorporation of the fine-grained spectral attention module, which effectively processes frequency domain information. Similarly, in the LQ version, D$^2$Fusion surpasses FADE \cite{tan2023deepfake} by up to 0.22 in ACC and AUC metrics. As compared to FADE, which relies solely on a multi-dependency graph in the spatial domain, our approach leverages the fine-grained spectral attention module, enabling more effective feature extraction and performance gains. However, in the HQ version, D$^2$Fusion exhibits slightly lower performance than IDM \cite{zhu2024deepfake}, with AUC decreased by 0.28. In the IDM model, it focuses on frequency domain information to decompose video frames into illumination and reflection components. This focus making it restrictive in very specific scenarios and thus resulting in suboptimal performance across different datasetst. In Table 6, when training on FF++(HQ) and testing on CD1, the AUC value of the IDM model decrease 11.6\% compared to our method. Besides, the discrepancy in D$^2$Fusion may be due to the loss of artifactual information in highly compressed videos, which also provides us with a future direction that the frequency domain might better preserve forged evidence in low-quality videos.

\subsubsection{Cross-manipulation Evaluation on FF++(HQ)}

In order to validate the generalization capability of our detection method, and demonstrate that our detection network is not restricted to detecting specific forgery generation techniques, we also evaluate the cross-manipulation performance of the proposed methods on FF++(HQ). We trained the involved models on one manipulation method and tested them on all methods with FF++(HQ). The result is shown in Table \ref{tab2}, obviously, the EfficientNet-B4 \cite{zhao2021multi}, MAT \cite{zhao2021multi} and GFF \cite{luo2021generalizing} work well in response to the known forgery operation. However, their ability to distinguish between authentic and fake images is limited when faced with unfamiliar forgery patterns. It is should also be noted that, compared to the CADDM \cite{dong2023implicit}, our method demonstrated considerable improvements in both intra- and cross-manipulations with NT, with an increase of AUC 4.05\%, 6.15\%, 2.04\%, and 0.34\% on the F2F, FS, DF, and NT, respectively. This task is particularly challenging due to the NT dataset uses GANs to generate fake videos that manipulate the mouth expressions, resulting in more realistic and localized artifact regions. Notably, our method achieved a remarkable 6.15\% improvement over the CADDM on the FS. It also outperformed the second-best method, DCL \cite{sun2022dual}, by 1.44\%, indicating the capability of $\mathrm{D^2}$Fusion to capture localized fake areas while addressing lower-level forgery manipulations. We also note that our proposed method exhibits slight performance regressions in certain scenarios, with a 3.52\% decrease against CADDM when trained with FS and tested in F2F, and a 2.68\% decrease when trained with DF and tested on FS. This could be attributed to an overrepresentation of artifact data type in the training set, which caused the model to overfit to a specific manipulation form. Overall, the results in Table 4 confirm that $\mathrm{D^2}$Fusion is effective in capturing artifact features across various forgery techniques by considering information from different domains.

\begin{table}[t]
\caption{Multi-source manipulation evaluation with other methods on FF++(HQ) in terms of $P$, $R$, $F1$, ACC, AUC (\%).} 
\resizebox{1\linewidth}{!}{\begin{tabular}{cccccccccccc}
\toprule
\multirow{2}{*}{Method} & \multirow{2}{*}{Year} & \multicolumn{5}{c}{DF(Auto-encoder)} & \multicolumn{5}{c}{F2F(Computer Graphics)} \\ \cmidrule(lr){3-7} \cmidrule(lr){8-12}
                        &                       & $P$ & $R$ & $F1$ & ACC & AUC & $P$  & $R$ & $F1$ & ACC & AUC \\ \cmidrule(lr){1-2} \cmidrule(lr){3-7} \cmidrule(lr){8-12}
MAT \cite{zhao2021multi}                  & 2021                  & 81.47 & 78.95 & 80.19 & 80.54 & 85.94 & 72.63 & 78.54 & 75.47 & 75.02 & 78.52  \\

 CADDM  \cite{dong2023implicit}                  & 2023                  & \underline{86.17} & \underline{89.97} & \underline{88.02} &\underline{87.79} & \underline{91.13} & 84.13 & 72.82 & 77.40 & 79.58 & 81.24 \\ 
 FreqNet    \cite{tan2024frequency}              & 2024                  & 84.38 & 86.66 & 85.50 & 85.33 & 89.75 &\textbf{ 87.81} & \underline{80.94} & \underline{84.24} & \underline{84.89} & \underline{86.32}\\
 \textbf{Ours}                     & 2024                  &\textbf{ 87.09} & \textbf{94.08} & \textbf{90.45} & \textbf{90.11} & \textbf{95.34} & \underline{87.25} & \textbf{82.35} & \textbf{84.73} & \textbf{85.70} & \textbf{87.23} \\
 \midrule
\multirow{2}{*}{Method} & \multirow{2}{*}{Year} & \multicolumn{5}{c}{FS(Computer Graphics)} & \multicolumn{5}{c}{NT(GAN)} \\ \cmidrule(lr){3-7} \cmidrule(lr){8-12}
                        &                       & $P$ & $R$ & $F1$ & ACC & AUC & $P$  & $R$ & $F1$ & ACC & AUC

\\ \cmidrule(lr){1-2} \cmidrule(lr){3-7} \cmidrule(lr){8-12}
MAT \cite{zhao2021multi}                    & 2021 & 74.70 & 62.19 & 67.87 & 70.63 & 72.56 & \underline{70.72} & 65.40 & 67.96 & 64.21 & 68.20 \\

 CADDM  \cite{dong2023implicit} &2023 & \underline{76.61} & 68.00 & 72.05 & 73.65 & 76.42 & 70.45 & \underline{70.31} & \underline{70.38} & \underline{70.45} & \underline{74.50} \\
 FreqNet    \cite{tan2024frequency}              & 2024    
& 74.19 & \textbf{78.43} & \textbf{76.30} & \underline{75.64} & \underline{80.23} & 66.90 & 67.90 & 67.40 & 67.22 & 69.94 \\
\textbf{Ours}                     & 2024     & \textbf{79.64} & \underline{71.01} & \underline{75.08} & \textbf{76.47} & \textbf{81.99} & \textbf{78.92} & \textbf{71.72} & \textbf{75.15} & \textbf{76.34} & \textbf{ 80.26 }\\
\bottomrule
\end{tabular}}
\label{multi-source}
\end{table}

\begin{table}[t]
\caption{Cross-dataset evaluation on CD1, CD2, DFDC and DFR by training FF++(HQ) compared with other methods in terms of AUC (\%).} 
\centering
\resizebox{0.5\linewidth}{!}{
\begin{tabular}{cccccc } \toprule
Method           &Year & CD1   & CD2   & DFDC  & DFR   \\ \midrule

Face X-ray \cite{li2020face}    &  2020 & 80.58 & 74.20 & 70.00 & -     \\
 MAT  \cite{zhao2021multi}  &2021 &  -     & 67.44 & 67.34 &   -    \\

 GFF \cite{luo2021generalizing}  &2021 &    -   & 76.65 & 71.58 & -      \\

TAN-GFD \cite{zhao2023tan}&2023 & - &72.33 &73.46& -\\

CADDM  \cite{dong2023implicit} &2023      &\textbf{89.57} & 77.04 & 71.49 & \underline{86.92} 
\\
 IDM \cite{zhu2024deepfake}&2024  &76.54& - &-&-\\
 E-TAD \cite{gao2024texture}& 2024&70.00 & 58.50&-
&-\\
 LSDA \cite{yan2024transcending}&2024 & 86.70 & \underline{83.00} & \underline{73.60}&-\\
 
 \textbf{Ours}  &2024 &\underline{88.14} & \textbf{83.29}  & \textbf{74.93}& \textbf{90.40} \\\bottomrule

\end{tabular}}
\label{cross_dataset}
\end{table}

\subsubsection{Multi-source Manipulation Evaluation}
Furthermore, we conduct a multi-source manipulation evaluation as an additional experiment to complement the cross-manipulation evaluation, given its inclusion of various manipulation operations. The performance of the proposed detection model under multi-source manipulation conditions is assessed on FF++(HQ). Specifically, the detection models are trained on three sub-datasets and tested on the remaining sub-dataset. As demonstrated in Table \ref{multi-source}, the proposed method significantly surpasses MAT \cite{zhao2021multi} in all the sub-datasets. Especially on the F2F dataset, the $P$ value of our method exceeds MAT by 14.62, since MAT only focuses on spatial information. Similar issues arise with CADDM \cite{dong2023implicit}; although it demonstrates relatively good generalization performance by extracting spatial features, its insufficient extraction of frequency domain information results in a 9.47\% reduction in the F1 score on the F2F dataset.  FreqNet \cite{tan2024frequency} performs well in F2F and FS datasets, which use computer graphics to generate Deepfake videos, but FreqNet underperforms against more advanced adversarial attacks such as those from Auto-encoders or GANs due to its bias towards specific high-level frequency information. Particularly, on the NT dataset, our method surpasses FreqNet by 12.02, 3.82, and 7.75 in $P$, $R$, $F1$ respectively.

\subsubsection{Cross-dataset Evaluation}
Although the aforementioned FF++ sub-datasets utilize four distinct forgery algorithms, the fake videos originate from identical source videos. To evaluate the performance of more different data distributions, we use the FF++(HQ) data set as the training data set and evaluate other data sets.
From Table \ref{cross_dataset}, it becomes evident that, when comparing our framework against various mainstream detection methods, our proposed $\mathrm{D^2}$Fusion significantly outperforms earlier basic reference methods, such as Face X-ray \cite{li2020face}. Compared to methods utilizing a single domain, such as MAT \cite{zhao2021multi} and GFF \cite{luo2021generalizing}, our approach demonstrates significant advantages. On the CD2 dataset, our method exceeds MAT by 23.50\%, and on the DFDC dataset, it surpasses GFF by 3.92\%. Notably, our method registers performance improvements of 8.11\% and 2.00\% over the advanced models TAN-GFD \cite{zhao2023tan} in CD2 and DFDC respectively. The AUC performance of our network exhibits a 4.00\% improvement over our baseline CADDM in DFR. Compared to most recent works \cite{zhu2024deepfake,gao2024texture}, our method outperforms E-TAD, IDM that only uses the frequency domain. For example, in CD1, our method exceeds E-TAD by 40.2\%; in DFDC, our method exceeds LSDA \cite{yan2024transcending} by 1.33. This improvement arises because our method delves deeper into feature exploration than LSDA. These results affirm that $\mathrm{D^2}$Fusion adeptly captures artifacts with greater precision, even amidst previously unseen backgrounds or identities.

In summary, our proposed $\mathrm{D^2}$Fusion outperforms previous SOTA techniques, showcasing its robustness and effectiveness in detecting forged content across a wide range of manipulation techniques and real-life scenarios.

\subsection{Qualitative Results}

The qualitative results are displayed in \Cref{qa}. We use the Equation (\ref{dissim}) to draw the DSSIM pictures to help intuitively identify tampered areas. Additionally, we employ Grad-CAM \cite{selvaraju2017grad} for generating heat maps, which visualize the focus regions of the models by fusing weights into the gradient information. In addition to vanilla ResNet34, we chose MAT \cite{zhao2021multi}, SBI \cite{shiohara2022detecting}, and CADDM \cite{dong2023implicit} as our comparison methods. To test whether these models are effective to capture artifact features between different forgery methods, we perform multi-source manipulation experiments results. 
\begin{figure*}[t]
\centering
\includegraphics[width=\linewidth]{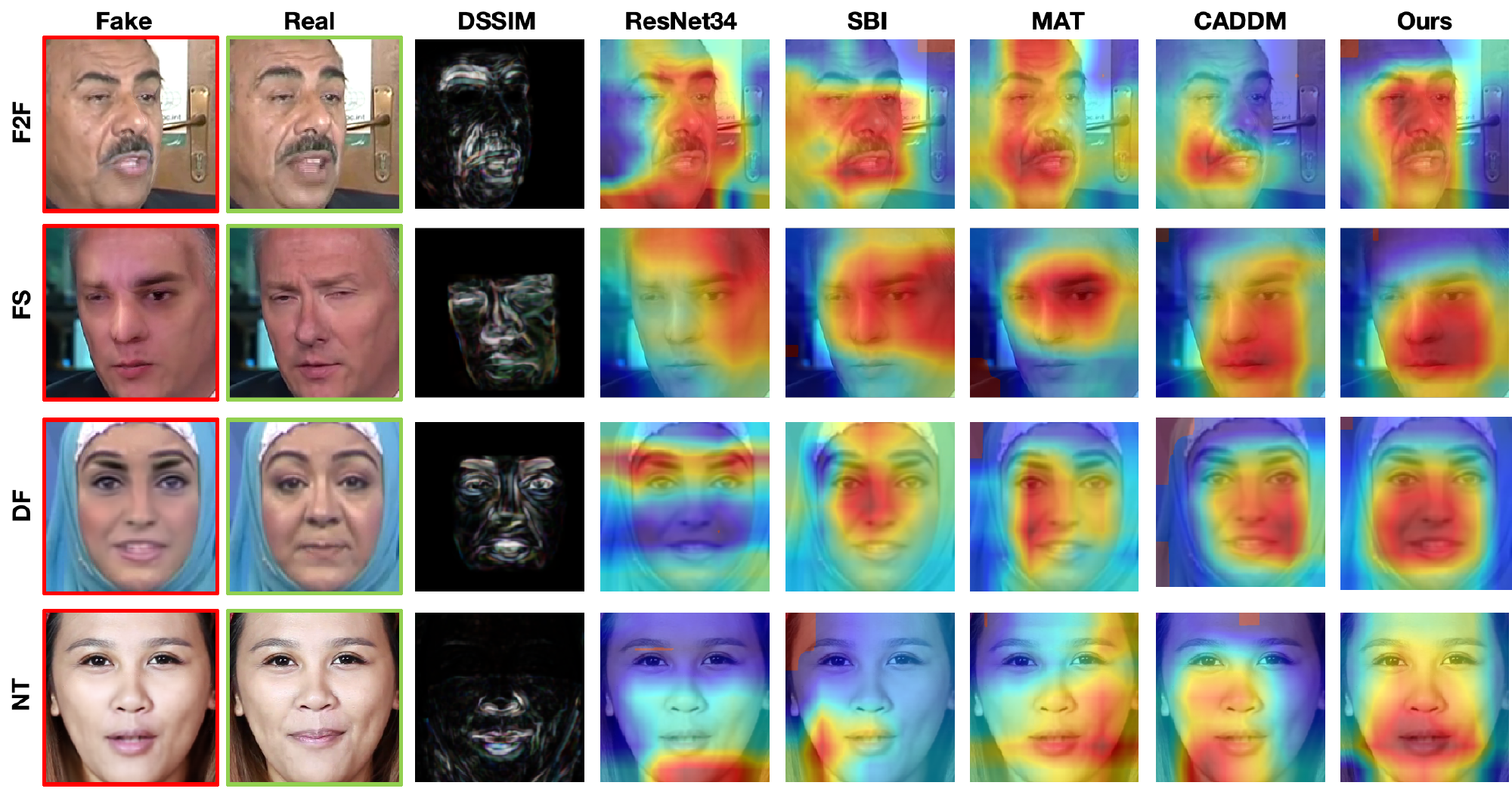}
\caption{Qualitative analysis on FF++(HQ). We show our detection results with multi-source manipulation evaluation. DSSIM images indicate the forgery area, and we can see that $\mathrm{D^2}$Fusion can better capture the manipulated area compared to other main-stream detection methods.}\label{qa}
\end{figure*}

From the heat maps of Resnet34, it becomes clear that this vanilla CNN are capable of detecting obvious blending boundary, which is a type of low-level feature easily perceptible even to the human eye. The detection performance is satisfactory for images with these evident clues in DF and FS, yet it proves ineffective for F2F and NT. Especially in NT, a facial reenactment dataset that alters expressions while preserving facial contours, ResNet34 still predominantly focuses on the boundary area.

SBI introduces self-blending images (the basis for multi-scale facial swap) for data augmentation, employing EfficientNet-B4 \cite{tan2019efficientnet} as the detection network. As depicted in Figure \ref{qa}, although SBI approximately identifies forgery areas, its accuracy is lacking. In F2F, FS, and NT, SBI inaccurately classifies some genuine areas as manipulated. In contrast to regular ResNet34, SBI enriches data diversity through self-blending, expanding detection beyond merely the edge regions. However, the inherent detection capabilities of the model remain limited.

In the case of MAT, the detection model employs multiple spatio attention heads. As illustrated in \Cref{qa}, the model is capable of focusing on various local parts. However, it fails to identify the general forgery area in these four example images accurately. Specifically, in NT, the area with the most significant mismatch between detected artifacts and actual forgeries is observed. This limitation stems from the exclusive reliance of MAT on spatial information, without incorporating data from other domains.

\begin{figure}[t]%
\centering
\includegraphics[width=0.7\textwidth]{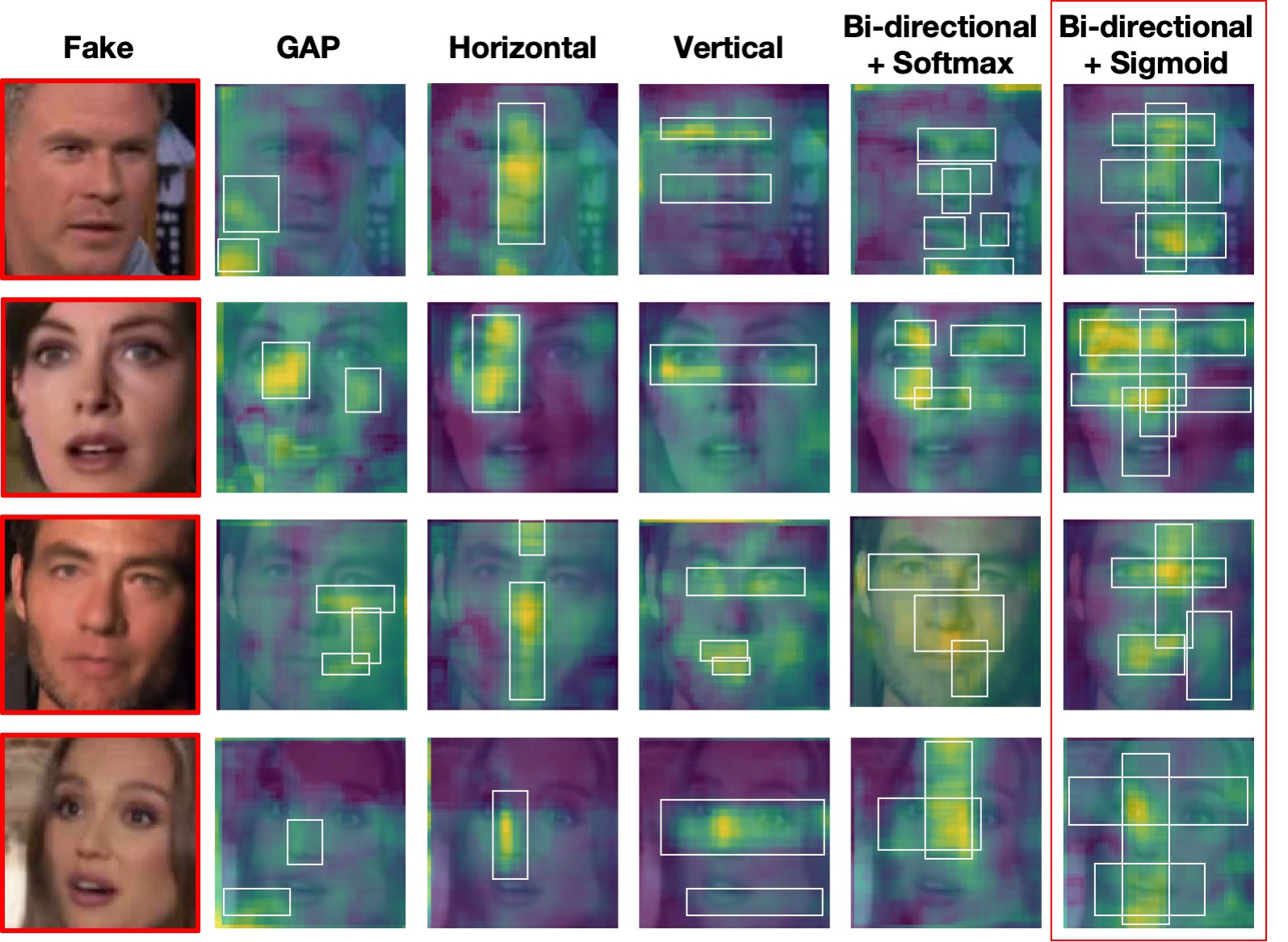}
\caption{From left to right are the fake images and the feature maps after GAP, horizontal average pooling, vertical average pooling, bi-directional average pooling with Softmax function and bi-directional average pooling with Sigmoid function. It should be noted that the forged features are marked by the white bounding boxes.}\label{bi_c}
\end{figure}

Compared to SBI and MAT, our baseline CADDM demonstrates precise alignment in detection, accurately pinpointing areas due to its utilization of multi-scale detection for local feature extraction. However, in F2F and DF, it does not capture all manipulated regions, suggesting that there is room for improvement in global feature learning. Our detection outcomes reveal that $\mathrm{D^2}$Fusion surpasses the baseline in effectiveness. In F2F, FS and DF, the localization closely matches the forged area, accurately encompassing a more extensive portion of the forgery compared to the baseline. In NT, the detection more accurately focuses on the forged mouth area. These results conclusively show that our model can efficiently identify artifacts, independent of the forgery techniques or whether the manipulation is global or localized.

\subsection{Ablation Studies}

\subsubsection{Evaluation on Components Intrinsic}

\begin{figure}[t]%
\centering
\includegraphics[width=0.7\textwidth]{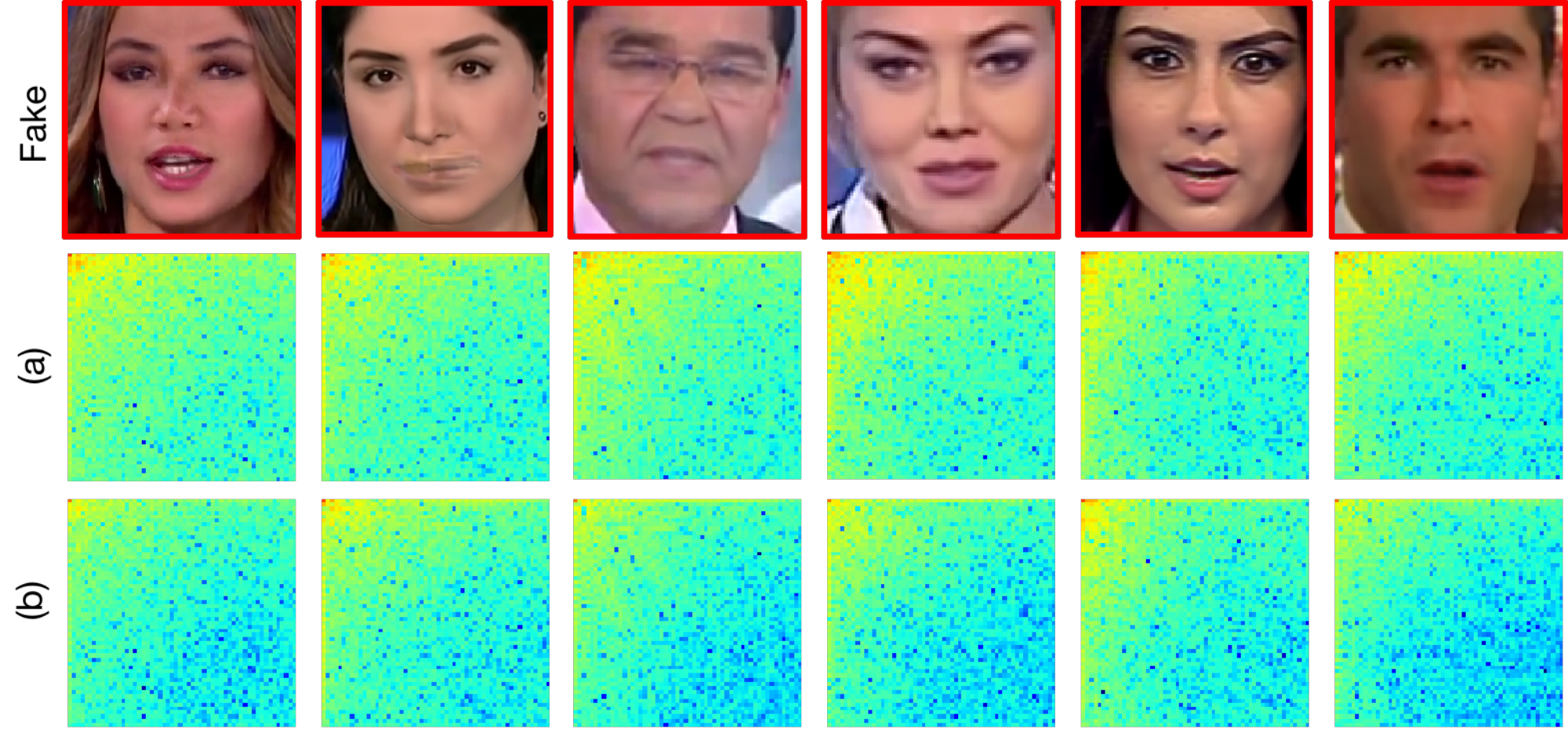}
\caption{(a) displays the log-scaled spectrum images after using a single DCT over the entire channel, (b) displays the log-scaled spectrum images after using multi-spectral partitioning and the respective corresponding DCTs.}\label{high}
\end{figure}

We particularly visualize the intrinsic effects of individual components on model performance. \Cref{bi_c} demonstrates that bi-directional average pooling can accurately localize manipulated regions compared to global average pooling (GAP). It can also effectively preserve the comprehensive spatial distribution of artifact clues from both directions compared to uni-directional average pooling. The visualization further reveals that using the Sigmoid function enables more concentrated and complete localization of artifact features. In contrast, while the Softmax function can also process features via bi-directional average pooling operations, its mapping results in a more dispersed localization of artifact features.

\begin{figure}[t]%
\centering
\includegraphics[width=0.8\textwidth]{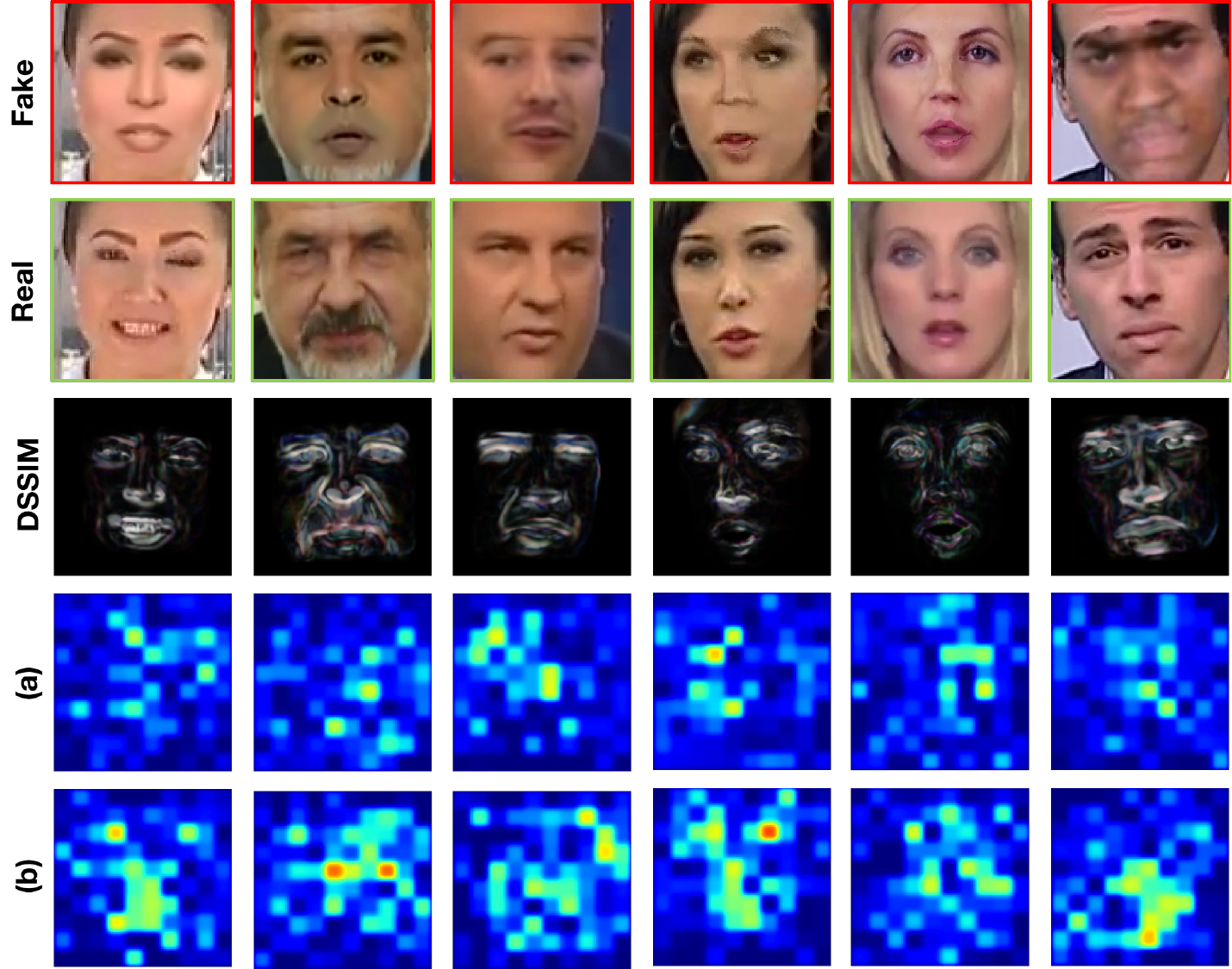}
\caption{(a) displays the error maps of $\vert{f_{f}-f_{r}}\vert$, in which $f_{f}$ is the fake image feature, $f_{r}$ is the real image feature. (b) displays the error maps of $\vert{f_{f}^{'}-f_{r}}\vert$, in which $f_{f}^{'}$ is the same fake image feature after feature superposition processing.}\label{error}
\end{figure}

\Cref{high} (a) illustrates that using a single DCT over the entire channel, the spectrum contains lower frequency components, while the spectrum band calculated from Equation (\ref{m_dct}) and Equation (\ref{c_s}) can retain stronger high-frequency components as shown in \Cref{high} (b). 

\Cref{error} shows the error maps before and after feature superposition. It is obvious that with the intervention of feature superposition, the difference between the authentic features and the artifact features within the image becomes more significant, which is manifested as a brighter region.

\begin{table}[]
\caption{Ablation study on the effect of different components of our model using five metrics (\%) in intra-dataset evaluation with FF++(HQ).} 
\resizebox{\linewidth}{!}{\begin{tabular}{cccccccc}
\toprule
\multirow{2}{*}{Bi-directional} & \multirow{2}{*}{Fine-grained Spectral} & \multirow{2}{*}{Feature Superposition} & \multicolumn{5}{c}{FF++} \\ \cline{4-8} 
                                &                                        &                                        & $P$  & $R$  & $F1$ & ACC & AUC \\ \cmidrule(lr){1-3}\cmidrule(lr){4-8}

                                  -     & - & - &97.23&95.09&96.14&96.20&97.25\\
 \Checkmark   & - & -  &   97.43&94.98&96.19& 96.22&98.79 \\      -  & \Checkmark  & -  &    96.02 & 96.79& 96.40& 96.41&98.45 \\     
     -  & -  & \Checkmark  &96.76&95.79&96.27&96.28&97.99\\
      - & \Checkmark & \Checkmark & 96.52 & \underline{97.29} &96.90 &96.88&98.99\\
  \Checkmark &           - & \Checkmark &96.17&98.40&97.27&97.24&99.13 \\
    \Checkmark &      \Checkmark     &  - & \underline{97.98}&\underline{97.29}&\underline{97.63}&\underline{97.65}&\underline{99.20}\\
  \Checkmark & \Checkmark & \Checkmark &  \textbf{98.08}&\textbf{97.39}&\textbf{97.73}&\textbf{97.77} &\textbf{99.42}
                          \\      \bottomrule
                          \label{abl_w}
\end{tabular}}
\end{table}

\subsubsection{Evaluation on Individual Component}

\Cref{abl_w} and \Cref{abl_c} reveal that the integration of a single domain module into vanilla ResNet34 proves to be effective. Specifically, as shown in \Cref{abl_w}, the addition of the fine-grained spectral module results in a 1.70 increase in $R$ for experiments within the dataset, indicating that the model is able to identify artifact samples more comprehensively. In cross-dataset experiments, DFR experiences the most significant improvement, with a 6.06\% rise in AUC. The incorporation of the fine-grained spectral attention module leads to AUC enhancements of 1.63\% in FF++, 2.10\% in CD1, 2.40\% in CD2, 3.07\% in DFDC, and 7.78\% in DFR. Implementing the feature superposition strategy yields $R$, $F1$, ACC, and AUC improvement of 0.70, 0.13, 0.08, 0.74 in intra-dataset scenarios, respectively, and a maximum increase AUC of 10.61\% in cross-dataset scenarios with DFR.

\begin{table}[t]
\caption{Ablation study on the effect of different components of our model using AUC (\%) metric in cross-dataset evaluation.} 
\begin{center}
    
\resizebox{\linewidth}{!}{
\begin{tabular}{ccccccc}
\toprule
Bi-directional & Fine-grained Spectral& Feature Superposition &CD1 & CD2 & DFDC & DFR\\ \cmidrule(lr){1-3} \cmidrule(lr){4-7}
  -     & - & -   & 83.43   &   79.27  & 67.34 &  69.50  \\
   \Checkmark   & - & -     & 84.65  & \underline{ 82.73 }  & 71.67 &  75.56  \\
  -    & \Checkmark & -  & 85.53    & 81.67 & 70.41   & 77.28\\
 -  &  - & \Checkmark &  84.37    & 82.72 & 69.20 & 80.11\\
  - & \Checkmark & \Checkmark & \underline{86.40}& 82.42 & 72.88 & 82.19 \\
 \Checkmark & - & \Checkmark&86.17 & 81.09 & 72.37 & \underline{87.42}\\

  \Checkmark & \Checkmark & -& 85.81 & 82.54 & \underline{73.46} & 85.80\\
  \Checkmark & \Checkmark & \Checkmark  &\textbf{88.14} &\textbf{83.29}&\textbf{73.93}&\textbf{90.40}\\ \bottomrule
\label{abl_c}
\end{tabular}
}
\end{center}
\end{table}

\begin{figure*}[t]
\centering
\includegraphics[width=\linewidth]{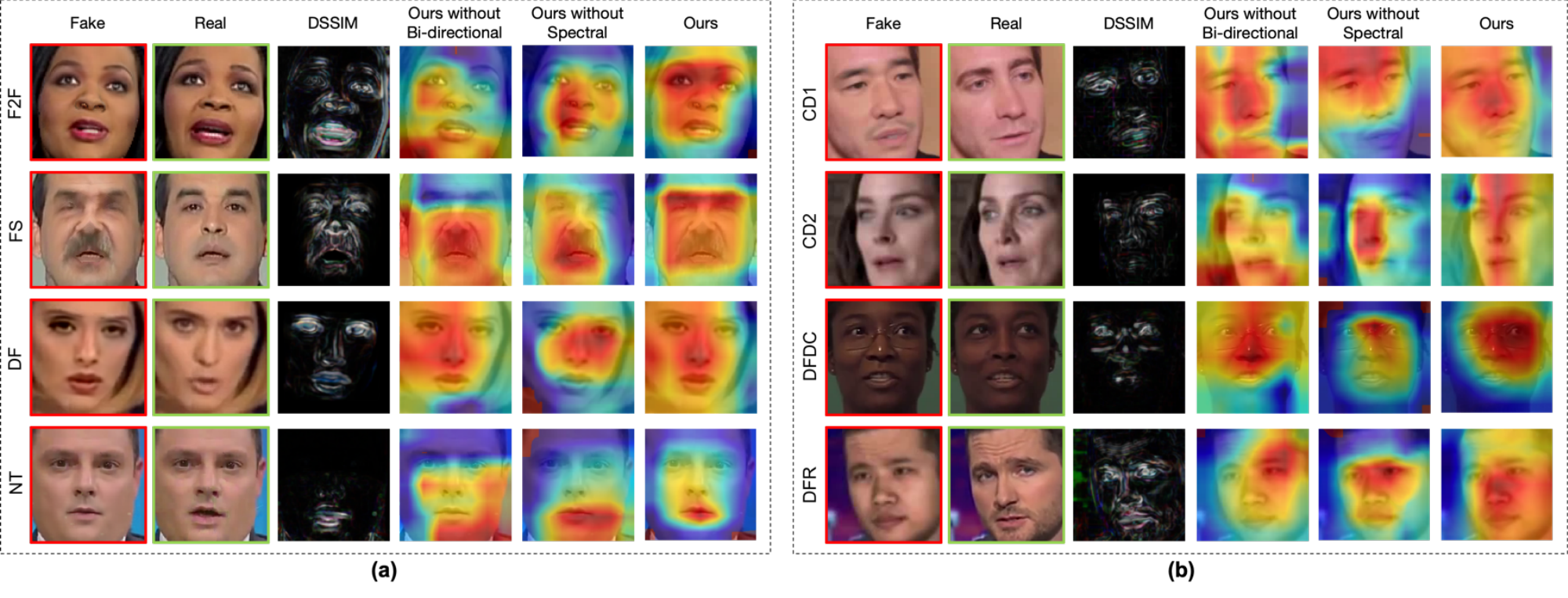}
\caption{The Grad-Cam visualizations of samples in different datasets. Sub-figure (a) exhibits intra-dataset experiments, and sub-figure (b) exhibits cross-dataset experiments. The heat maps prove that with the introduction of our attention modules for different forgery methods and real-word information, $\mathrm{D^2}$Fusion can more precisely capture the forgery region.}\label{heat}
\end{figure*}

We utilize visualization tools to highlight the critical role of our designed attention modules. As depicted in \Cref{heat} (a) for the intra-dataset scenario, it becomes evident that models lacking the bi-directional module can identify tampered areas, yet with insufficient precision, particularly in FS. This situation suggests that spatial information could be more effectively processed. Conversely, models missing the fine-grained spectral attention module might accurately locate tampered zones but fail to encompass all manipulated areas as comprehensively as desired. Such findings indicate the underutilization of frequency domain information, signalling opportunities for enhancement. In the heat maps of our model, it is observed that the attention mechanisms enable the model to localize the manipulation area with greater precision, regardless of the manipulation technique employed.  

In Figure \ref{heat} (b), we present visualizations for cross-dataset scenarios. The outcomes from models lacking bi-directional attention reveal instances of positioning overflow, notably in CD1 and CD2. Similarly, models devoid of fine-grained spectral attention tend to identify smaller forgery areas across datasets. In stark contrast, our comprehensive network showcases remarkable robustness, precisely locating and identifying manipulated regions, regardless of the complexity of real-world scenarios or the diversity of ID representations. This demonstrates that adding the attention components enhances the ability of the model to capture the abnormal regions and significantly improves its adaptability in dealing with variable and unknown pattern modifications. This phenomenon also accords with our motivation. 

\begin{figure*}[t]
\centering
\includegraphics[width=\linewidth]{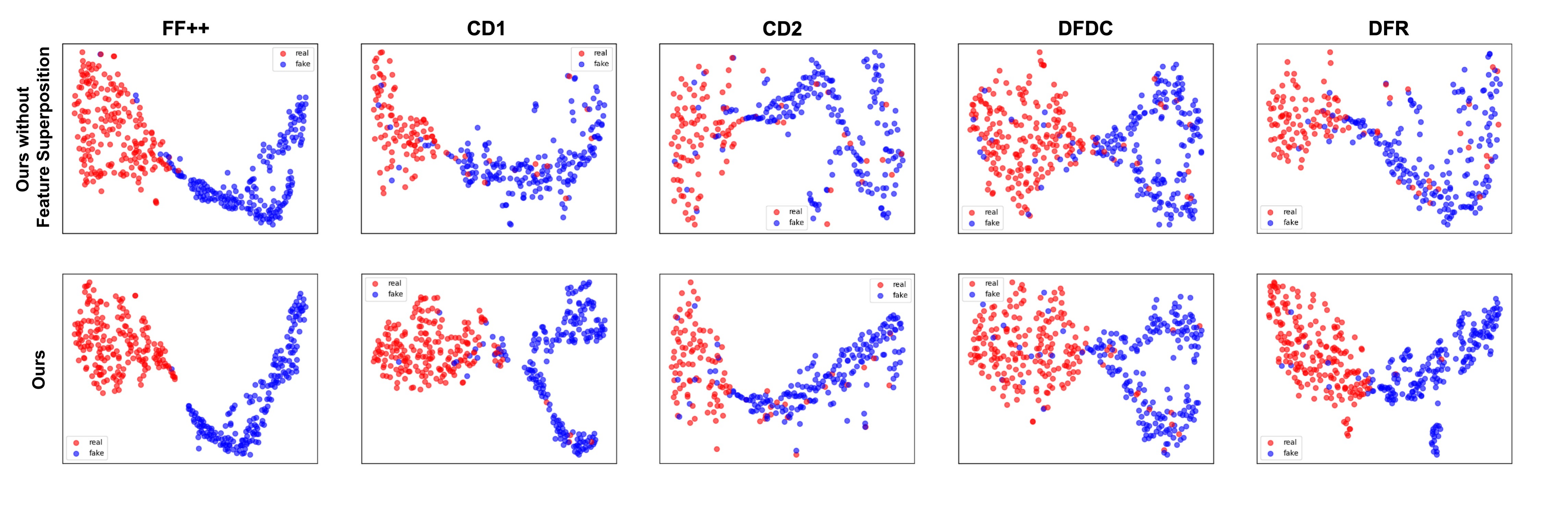}
\caption{Visualization of t-SNE feature embedding in different scenarios, note that the red dots represent the real samples, while the blue dots represent the fake samples.}\label{tsne}
\end{figure*}

Finally, we apply t-SNE \cite{van2008visualizing} to visualize feature vector distribution in testing dataset from the last layers of the models. The visual results, as depicted in Figure \ref{tsne}, reveal that the feature distribution boundary appears relatively unambiguous in models lacking the feature superposition step, especially in CD1 and CD2. In contrast, the different categories of samples after the feature superposition process are further separated in the t-SNE embedding space, thus reflecting that introducing our feature superposition can help the model better increase the clarity of decision-making boundaries. This phenomenon is also reflected in Table \ref{abl_c}, where within our complete framework, the introduction of feature superposition causes the AUC to rise by 0.09\% in FF++(HQ), 2.33\% in CD1, 0.75\% in CD2, 0.47\% in DFDC, 4.60\% in DFR.

\begin{table}[t]
\caption{Network overload comparison in terms of Parameters(M) and FLOPs(G) with ACC in intra-dataset evaluation on FF++(HQ). In () represents the difference compared to the corresponding backbone.} 
\begin{center}
\resizebox{0.8\linewidth}{!}{\begin{tabular}{ccccc} \toprule
Model                         & Parameters(M) & FLOPs(G)&ACC\\ \midrule

EfficientNetB4   \cite{koonce2021efficientnet}             &    33.48           &1.60  &96.13    \\

MAT(EfficientNetB4-based)  \cite{zhao2021multi}  & 39.44(\underline{5.96$\uparrow$})  &2.17(\underline{0.57$\uparrow$}) & 97.60(\underline{1.47$\uparrow$})\\

TAN-GFD(EfficientNetB4-based) \cite{zhao2023tan} &       46.87(13.39$\uparrow$)  &4.34(2.74$\uparrow$) &97.17(1.04$\uparrow$)               \\
\midrule
Xception \cite{chollet2017xception}   &  39.70     
& 4.61&95.49\\
GFF (Xception-based) \cite{luo2021generalizing} & 53.25(13.55$\uparrow$)& 10.50(5.89$\uparrow$) &96.87(1.38$\uparrow$)\\ \midrule
ResNet34  \cite{he2016deep}   &   37.19  & 3.68 &96.20      \\

CADDM(ResNet34-based) \cite{dong2023implicit}        &      40.60(\textbf{3.41$\uparrow$})         & 4.08(\textbf{0.40$\uparrow$})     &\underline{97.59}(1.39$\uparrow$)  \\
\textbf{Ours(ResNet34-based)}          &      45.66(8.47$\uparrow$)     & 5.97(2.29$\uparrow$)   & \textbf{97.77}(\textbf{1.57$\uparrow$})  \\ \bottomrule
\end{tabular}}
\label{parameters}
\end{center}
\end{table}
\subsection{Evaluation of Network Load}
To illustrate network load, \Cref{parameters} presents the total number of training parameters, which measure the size of models, and Floating Point Operations per Second (FLOPs) to assess the complexity of models. In detection within the FF++(HQ) dataset, D$^2$Fusion outperforms all approaches in \Cref{parameters} in terms of accuracy with a relatively small network overload increase. 

It is evident that baseline networks such as EfficientNetB4 \cite{koonce2021efficientnet}, Xception \cite{chollet2017xception}, and ResNet34 \cite{he2016deep} have fewer parameters and FLOPs because these networks serve solely as classification models without any architectural reconfiguration or additional modules tailored. This limitation results in poorer detection performance in deepfake detection tasks. In networks using EfficientNetB4 as the backbone, the MAT \cite{zhao2021multi}, and TAN-GFD \cite{zhao2023tan} models see an increase in parameters by 5.96M and 13.39M, respectively. The former arises from introduced attention modules, while the latter incorporates the processing of multi-scale texture difference features. In networks with Xception as the backbone, GFF  \cite{luo2021generalizing} significantly increases 5.89G in FLOPs due to its dual-stream architecture designed to process both spatio-frequency features. However, although D$^2$Fusion also addresses spatio-frequency features and has more backbone parameters than those in GFF, the overall parameter for our model is substantially less than that of GFF. Compared to CADDM, D$^2$Fusion significantly enhances generalization performance with a relatively small network overload increase of 1.89G more FLOPs and exceeds CADDM by 0.18 ACC. For its backbone ResNet34, although D$^2$Fusion incurs an additional 2.29G FLOPs, it achieves the most significant increase in accuracy, with an improvement of 1.57.

\begin{figure}[]%
\centering
\includegraphics[width=1\textwidth]{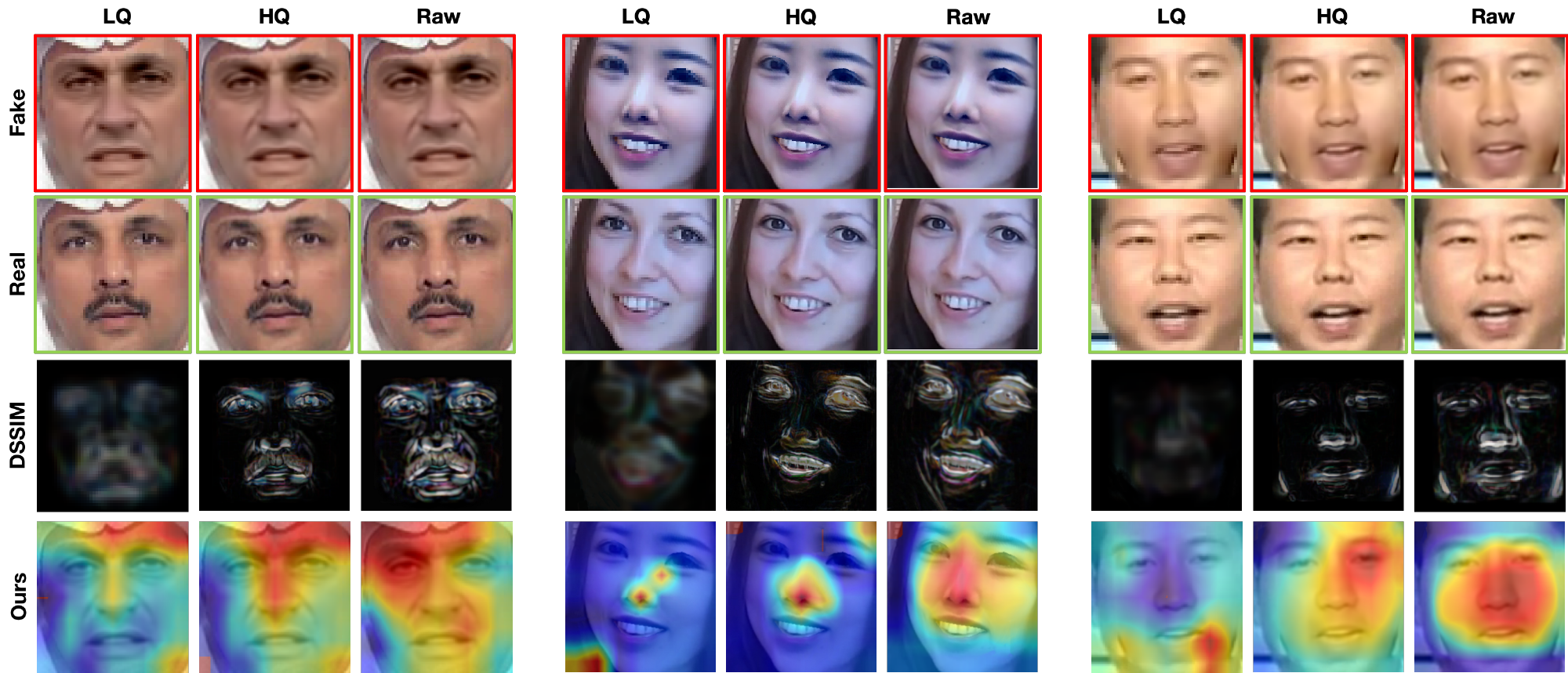}
\caption{Failure cases. 
These samples all fail to be detected under low-quality video conditions, but they are successfully detected in high-quality and raw videos.}\label{failure}
\end{figure}

\section{Limitation} \label{limitation}
As \Cref{tab1} demonstrates, our method exhibits a decrease in performance when dealing with high compression rate videos, in other words, low-quality videos. The primary cause of this decline is the high compression rate, causing the loss of forgery clues within the video. \Cref{failure} clearly illustrates the impact of video compression rate on detection outcomes. In \Cref{failure}, the human eyes may struggle to differentiate images at varying compression rates, but the DSSIM visualization demonstrates how an increased compression rate results in a decline in the loss of artifact information. Consequently, D$^2$Fusion erroneously focuses on areas unrelated to the forged regions during low-quality video detection. As the compression rate declines and video quality improves, the increase in forgery details prompts the model to adjust its area of focus, thereby correcting the detection outcomes.

\section{Future work}

As discussed in \Cref{limitation}, video quality can interfere with detection outcomes, and in real life, detectors often encounter batches of low-quality videos, due to the varying levels of compression applied during the transmission through social media networks. Therefore, detectors need to enhance their ability to handle high-compression rate videos in the future. To this end, the following optimization directions are proposed:
\begin{compactitem}

\item {We will persistently refine and deepen our research on capturing local positional information of artifact clues from the spatial domain. This effort aims to achieve more precise localization and extend our research findings to practical applications.}
\item {We plan to explore the impact of various frequency compression methods, such as wavelet transforms, on the preservation of forgery clues. Through this investigation, we aim to gain a deeper understanding of the effectiveness of different frequency compression technologies in maintaining original forgery features.}
\item {We plan to incorporate recent datasets focused on highly compressed videos to explore low-quality video detection. This direction aims to adapt to the properties of low-quality videos, thereby enhancing the performance and reliability of detectors in practical media.}
    
\end{compactitem}

\section{Conclusion}
Existing works do not take into account the local and global properties of Deepfake videos in different domains. They also lack consideration for addressing the fundamental interaction issues in between different domains by utilizing the local and global complementarity in the spatial and frequency domain, respectively. To address the problem of insufficient feature processing and interaction in Deepfake detection, in this work, we introduce a novel facial manipulation detection network, termed D$^2$Fusion. Specifically, we first introduce a bi-directional attention module designed to average features horizontally and vertically. This module effectively maintains the spatial distribution of artifacts, thus improving the precision of artifact localization. To better extract sufficiently detailed artifact information, we utilize a fine-grained frequency attention module. By preserving the high-frequency components, this module significantly captures more of the details in the artifact features. In addition, our designed feature superposition strategy accepts two domain features to amplify the difference between authentic and artifact features, thus helping the detection model to be more generalized across different manipulation operations and real-world scenarios. 

We evaluate our detection network in various experimental scenarios with significantly enhanced detection capabilities. In intra-dataset evaluation, D$^2$Fusion exceeds CADDM \cite{dong2023implicit} by 0.18 ACC within FF++(HQ). Furthermore, the model exhibits strong generalization capabilities in cross-dataset evaluations. 
On CD1, our method surpasses the emerging LSDA \cite{yan2024transcending} by 1.44 in AUC, and this is critical for practical applications where the model must perform well on previously unseen techniques and real-life scenarios. 

However, our work acknowledges limitations in the performance of D$^2$Fusion model under high video compression. In the future, we propose further enhancements for handling such videos by improving local artifact detection and different frequency transforms \cite{duan2023dynamic,miao2025laser} with advanced low-quality video datasets. Our work aims to enhance deepfake detection techniques for practical implementation, especially targeting the identification of manipulated content in low-quality videos frequently transmitted through social media platforms.




\bibliographystyle{elsarticle-num-names} 
\bibliography{new_bibliography.bib}

\begin{thebibliography}{59}
\expandafter\ifx\csname natexlab\endcsname\relax\def\natexlab#1{#1}\fi
\providecommand{\url}[1]{\texttt{#1}}
\providecommand{\href}[2]{#2}
\providecommand{\path}[1]{#1}
\providecommand{\DOIprefix}{doi:}
\providecommand{\ArXivprefix}{arXiv:}
\providecommand{\URLprefix}{URL: }
\providecommand{\Pubmedprefix}{pmid:}
\providecommand{\doi}[1]{\href{http://dx.doi.org/#1}{\path{#1}}}
\providecommand{\Pubmed}[1]{\href{pmid:#1}{\path{#1}}}
\providecommand{\bibinfo}[2]{#2}
\ifx\xfnm\relax \def\xfnm[#1]{\unskip,\space#1}\fi
\bibitem[{Yadav and Salmani(2019)}]{yadav2019deepfake}
\bibinfo{author}{D.~Yadav}, \bibinfo{author}{S.~Salmani},
\newblock \bibinfo{title}{Deepfake: A survey on facial forgery technique using generative adversarial network},
\newblock in: \bibinfo{booktitle}{2019 International conference on intelligent computing and control systems}, \bibinfo{organization}{IEEE}, \bibinfo{year}{2019}, pp. \bibinfo{pages}{852--857}.
\bibitem[{Tolosana et~al.(2020)Tolosana, Vera-Rodriguez, Fierrez, Morales, and Ortega-Garcia}]{tolosana2020deepfakes}
\bibinfo{author}{R.~Tolosana}, \bibinfo{author}{R.~Vera-Rodriguez}, \bibinfo{author}{J.~Fierrez}, \bibinfo{author}{A.~Morales}, \bibinfo{author}{J.~Ortega-Garcia},
\newblock \bibinfo{title}{Deepfakes and beyond: A survey of face manipulation and fake detection},
\newblock \bibinfo{journal}{Information Fusion} \bibinfo{volume}{64} (\bibinfo{year}{2020}) \bibinfo{pages}{131--148}.
\bibitem[{Chollet(2017)}]{chollet2017xception}
\bibinfo{author}{F.~Chollet},
\newblock \bibinfo{title}{Xception: Deep learning with depthwise separable convolutions},
\newblock in: \bibinfo{booktitle}{Proceedings of the IEEE/CVF conference on computer vision and pattern recognition}, \bibinfo{year}{2017}, pp. \bibinfo{pages}{1251--1258}.
\bibitem[{He et~al.(2016)He, Zhang, Ren, and Sun}]{he2016deep}
\bibinfo{author}{K.~He}, \bibinfo{author}{X.~Zhang}, \bibinfo{author}{S.~Ren}, \bibinfo{author}{J.~Sun},
\newblock \bibinfo{title}{Deep residual learning for image recognition},
\newblock in: \bibinfo{booktitle}{Proceedings of the IEEE/CVF conference on computer vision and pattern recognition}, \bibinfo{year}{2016}, pp. \bibinfo{pages}{770--778}.
\bibitem[{Tan and Le(2019)}]{tan2019efficientnet}
\bibinfo{author}{M.~Tan}, \bibinfo{author}{Q.~Le},
\newblock \bibinfo{title}{Efficientnet: Rethinking model scaling for convolutional neural networks},
\newblock in: \bibinfo{booktitle}{International conference on machine learning}, \bibinfo{organization}{PMLR}, \bibinfo{year}{2019}, pp. \bibinfo{pages}{6105--6114}.
\bibitem[{Li et~al.(2020)Li, Bao, Zhang, Yang, Chen, Wen, and Guo}]{li2020face}
\bibinfo{author}{L.~Li}, \bibinfo{author}{J.~Bao}, \bibinfo{author}{T.~Zhang}, \bibinfo{author}{H.~Yang}, \bibinfo{author}{D.~Chen}, \bibinfo{author}{F.~Wen}, \bibinfo{author}{B.~Guo},
\newblock \bibinfo{title}{Face x-ray for more general face forgery detection},
\newblock in: \bibinfo{booktitle}{Proceedings of the IEEE/CVF conference on computer vision and pattern recognition}, \bibinfo{year}{2020}, pp. \bibinfo{pages}{5001--5010}.
\bibitem[{Shiohara and Yamasaki(2022)}]{shiohara2022detecting}
\bibinfo{author}{K.~Shiohara}, \bibinfo{author}{T.~Yamasaki},
\newblock \bibinfo{title}{Detecting deepfakes with self-blended images},
\newblock in: \bibinfo{booktitle}{Proceedings of the IEEE/CVF conference on computer vision and pattern recognition}, \bibinfo{year}{2022}, pp. \bibinfo{pages}{18720--18729}.
\bibitem[{Yu et~al.(2023)Yu, Li, Yang, Li, Li, and Zhang}]{yu2023fdml}
\bibinfo{author}{M.~Yu}, \bibinfo{author}{H.~Li}, \bibinfo{author}{J.~Yang}, \bibinfo{author}{X.~Li}, \bibinfo{author}{S.~Li}, \bibinfo{author}{J.~Zhang},
\newblock \bibinfo{title}{Fdml: Feature disentangling and multi-view learning for face forgery detection},
\newblock \bibinfo{journal}{Neurocomputing}  (\bibinfo{year}{2023}) \bibinfo{pages}{127192}.
\bibitem[{Durall et~al.(2019)Durall, Keuper, Pfreundt, and Keuper}]{durall2019unmasking}
\bibinfo{author}{R.~Durall}, \bibinfo{author}{M.~Keuper}, \bibinfo{author}{F.-J. Pfreundt}, \bibinfo{author}{J.~Keuper},
\newblock \bibinfo{title}{Unmasking deepfakes with simple features},
\newblock \bibinfo{journal}{arXiv:1911.00686}  (\bibinfo{year}{2019}).
\bibitem[{Frank et~al.(2020)Frank, Eisenhofer, Sch{\"o}nherr, Fischer, Kolossa, and Holz}]{frank2020leveraging}
\bibinfo{author}{J.~Frank}, \bibinfo{author}{T.~Eisenhofer}, \bibinfo{author}{L.~Sch{\"o}nherr}, \bibinfo{author}{A.~Fischer}, \bibinfo{author}{D.~Kolossa}, \bibinfo{author}{T.~Holz},
\newblock \bibinfo{title}{Leveraging frequency analysis for deep fake image recognition},
\newblock in: \bibinfo{booktitle}{International conference on machine learning}, \bibinfo{organization}{PMLR}, \bibinfo{year}{2020}, pp. \bibinfo{pages}{3247--3258}.
\bibitem[{Jeong et~al.(2022)Jeong, Kim, Ro, and Choi}]{jeong2022frepgan}
\bibinfo{author}{Y.~Jeong}, \bibinfo{author}{D.~Kim}, \bibinfo{author}{Y.~Ro}, \bibinfo{author}{J.~Choi},
\newblock \bibinfo{title}{Frepgan: robust deepfake detection using frequency-level perturbations},
\newblock in: \bibinfo{booktitle}{Proceedings of the AAAI conference on artificial intelligence}, volume~\bibinfo{volume}{36}, \bibinfo{year}{2022}, pp. \bibinfo{pages}{1060--1068}.
\bibitem[{Agarwal et~al.(2021)Agarwal, Agarwal, Sinha, Vatsa, and Singh}]{agarwal2021md}
\bibinfo{author}{A.~Agarwal}, \bibinfo{author}{A.~Agarwal}, \bibinfo{author}{S.~Sinha}, \bibinfo{author}{M.~Vatsa}, \bibinfo{author}{R.~Singh},
\newblock \bibinfo{title}{Md-csdnetwork: Multi-domain cross stitched network for deepfake detection},
\newblock in: \bibinfo{booktitle}{2021 16th IEEE international conference on automatic face and gesture recognition (FG 2021)}, \bibinfo{organization}{IEEE}, \bibinfo{year}{2021}, pp. \bibinfo{pages}{1--8}.
\bibitem[{Wang et~al.(2023)Wang, Yu, Chen, Hu, and Peng}]{wang2023dynamic}
\bibinfo{author}{Y.~Wang}, \bibinfo{author}{K.~Yu}, \bibinfo{author}{C.~Chen}, \bibinfo{author}{X.~Hu}, \bibinfo{author}{S.~Peng},
\newblock \bibinfo{title}{Dynamic graph learning with content-guided spatial-frequency relation reasoning for deepfake detection},
\newblock in: \bibinfo{booktitle}{Proceedings of the IEEE/CVF conference on computer vision and pattern recognition}, \bibinfo{year}{2023}, pp. \bibinfo{pages}{7278--7287}.
\bibitem[{Miao et~al.(2023)Miao, Tan, Chu, Liu, Hu, and Yu}]{miao2023f}
\bibinfo{author}{C.~Miao}, \bibinfo{author}{Z.~Tan}, \bibinfo{author}{Q.~Chu}, \bibinfo{author}{H.~Liu}, \bibinfo{author}{H.~Hu}, \bibinfo{author}{N.~Yu},
\newblock \bibinfo{title}{F 2 trans: High-frequency fine-grained transformer for face forgery detection},
\newblock \bibinfo{journal}{IEEE Transactions on Information Forensics and Security} \bibinfo{volume}{18} (\bibinfo{year}{2023}) \bibinfo{pages}{1039--1051}.
\bibitem[{Yu et~al.(2021)Yu, Xia, Fei, and Lu}]{yu2021survey}
\bibinfo{author}{P.~Yu}, \bibinfo{author}{Z.~Xia}, \bibinfo{author}{J.~Fei}, \bibinfo{author}{Y.~Lu},
\newblock \bibinfo{title}{A survey on deepfake video detection},
\newblock \bibinfo{journal}{Iet Biometrics} \bibinfo{volume}{10} (\bibinfo{year}{2021}) \bibinfo{pages}{607--624}.
\bibitem[{Lin et~al.(2012)Lin, Lin, Tang, and Wang}]{lin2012face}
\bibinfo{author}{Y.~Lin}, \bibinfo{author}{Q.~Lin}, \bibinfo{author}{F.~Tang}, \bibinfo{author}{S.~Wang},
\newblock \bibinfo{title}{Face replacement with large-pose differences},
\newblock in: \bibinfo{booktitle}{Proceedings of the 20th ACM international conference on Multimedia}, \bibinfo{year}{2012}, pp. \bibinfo{pages}{1249--1250}.
\bibitem[{Nirkin et~al.(2018)Nirkin, Masi, Tuan, Hassner, and Medioni}]{nirkin2018face}
\bibinfo{author}{Y.~Nirkin}, \bibinfo{author}{I.~Masi}, \bibinfo{author}{A.~T. Tuan}, \bibinfo{author}{T.~Hassner}, \bibinfo{author}{G.~Medioni},
\newblock \bibinfo{title}{On face segmentation, face swapping, and face perception},
\newblock in: \bibinfo{booktitle}{2018 13th IEEE International Conference on Automatic Face \& Gesture Recognition (FG 2018)}, \bibinfo{organization}{IEEE}, \bibinfo{year}{2018}, pp. \bibinfo{pages}{98--105}.
\bibitem[{Smith and Zhang(2012)}]{smith2012joint}
\bibinfo{author}{B.~M. Smith}, \bibinfo{author}{L.~Zhang},
\newblock \bibinfo{title}{Joint face alignment with non-parametric shape models},
\newblock in: \bibinfo{booktitle}{Computer Vision--ECCV 2012: 12th European conference on computer vision, Florence, Italy, October 7-13, 2012, Proceedings, Part III 12}, \bibinfo{organization}{Springer}, \bibinfo{year}{2012}, pp. \bibinfo{pages}{43--56}.
\bibitem[{Rana et~al.(2022)Rana, Nobi, Murali, and Sung}]{rana2022deepfake}
\bibinfo{author}{M.~S. Rana}, \bibinfo{author}{M.~N. Nobi}, \bibinfo{author}{B.~Murali}, \bibinfo{author}{A.~H. Sung},
\newblock \bibinfo{title}{Deepfake detection: A systematic literature review},
\newblock \bibinfo{journal}{IEEE access} \bibinfo{volume}{10} (\bibinfo{year}{2022}) \bibinfo{pages}{25494--25513}.
\bibitem[{Li et~al.(2019)Li, Bao, Yang, Chen, and Wen}]{li2019faceshifter}
\bibinfo{author}{L.~Li}, \bibinfo{author}{J.~Bao}, \bibinfo{author}{H.~Yang}, \bibinfo{author}{D.~Chen}, \bibinfo{author}{F.~Wen},
\newblock \bibinfo{title}{Faceshifter: Towards high fidelity and occlusion aware face swapping},
\newblock \bibinfo{journal}{arXiv preprint arXiv:1912.13457}  (\bibinfo{year}{2019}).
\bibitem[{Perov et~al.(2020)Perov, Gao, Chervoniy, Liu, Marangonda, Um{\'e}, Dpfks, Facenheim, RP, Jiang et~al.}]{perov2020deepfacelab}
\bibinfo{author}{I.~Perov}, \bibinfo{author}{D.~Gao}, \bibinfo{author}{N.~Chervoniy}, \bibinfo{author}{K.~Liu}, \bibinfo{author}{S.~Marangonda}, \bibinfo{author}{C.~Um{\'e}}, \bibinfo{author}{M.~Dpfks}, \bibinfo{author}{C.~S. Facenheim}, \bibinfo{author}{L.~RP}, \bibinfo{author}{J.~Jiang}, et~al.,
\newblock \bibinfo{title}{Deepfacelab: Integrated, flexible and extensible face-swapping framework},
\newblock \bibinfo{journal}{arXiv preprint arXiv:2005.05535}  (\bibinfo{year}{2020}).
\bibitem[{Kingma and Welling(2013)}]{kingma2013auto}
\bibinfo{author}{D.~P. Kingma}, \bibinfo{author}{M.~Welling},
\newblock \bibinfo{title}{Auto-encoding variational bayes},
\newblock \bibinfo{journal}{arXiv preprint arXiv:1312.6114}  (\bibinfo{year}{2013}).
\bibitem[{Choi et~al.(2018)Choi, Choi, Kim, Ha, Kim, and Choo}]{choi2018stargan}
\bibinfo{author}{Y.~Choi}, \bibinfo{author}{M.~Choi}, \bibinfo{author}{M.~Kim}, \bibinfo{author}{J.-W. Ha}, \bibinfo{author}{S.~Kim}, \bibinfo{author}{J.~Choo},
\newblock \bibinfo{title}{Stargan: Unified generative adversarial networks for multi-domain image-to-image translation},
\newblock in: \bibinfo{booktitle}{Proceedings of the IEEE/CVF conference on computer vision and pattern recognition}, \bibinfo{year}{2018}, pp. \bibinfo{pages}{8789--8797}.
\bibitem[{Karras et~al.(2020)Karras, Laine, Aittala, Hellsten, Lehtinen, and Aila}]{karras2020analyzing}
\bibinfo{author}{T.~Karras}, \bibinfo{author}{S.~Laine}, \bibinfo{author}{M.~Aittala}, \bibinfo{author}{J.~Hellsten}, \bibinfo{author}{J.~Lehtinen}, \bibinfo{author}{T.~Aila},
\newblock \bibinfo{title}{Analyzing and improving the image quality of stylegan},
\newblock in: \bibinfo{booktitle}{Proceedings of the IEEE/CVF conference on computer vision and pattern recognition}, \bibinfo{year}{2020}, pp. \bibinfo{pages}{8110--8119}.
\bibitem[{Nirkin et~al.(2019)Nirkin, Keller, and Hassner}]{nirkin2019fsgan}
\bibinfo{author}{Y.~Nirkin}, \bibinfo{author}{Y.~Keller}, \bibinfo{author}{T.~Hassner},
\newblock \bibinfo{title}{Fsgan: Subject agnostic face swapping and reenactment},
\newblock in: \bibinfo{booktitle}{Proceedings of the IEEE/CVF international conference on computer vision}, \bibinfo{year}{2019}, pp. \bibinfo{pages}{7184--7193}.
\bibitem[{Li and Lyu(2018)}]{li2018exposing}
\bibinfo{author}{Y.~Li}, \bibinfo{author}{S.~Lyu},
\newblock \bibinfo{title}{Exposing deepfake videos by detecting face warping artifacts},
\newblock \bibinfo{journal}{arXiv preprint arXiv:1811.00656}  (\bibinfo{year}{2018}).
\bibitem[{Cozzolino et~al.(2021)Cozzolino, R{\"o}ssler, Thies, Nie{\ss}ner, and Verdoliva}]{cozzolino2021id}
\bibinfo{author}{D.~Cozzolino}, \bibinfo{author}{A.~R{\"o}ssler}, \bibinfo{author}{J.~Thies}, \bibinfo{author}{M.~Nie{\ss}ner}, \bibinfo{author}{L.~Verdoliva},
\newblock \bibinfo{title}{Id-reveal: Identity-aware deepfake video detection},
\newblock in: \bibinfo{booktitle}{Proceedings of the IEEE/CVF international conference on computer vision}, \bibinfo{year}{2021}, pp. \bibinfo{pages}{15108--15117}.
\bibitem[{Dong et~al.(2022)Dong, Bao, Chen, Zhang, Zhang, Yu, Chen, Wen, and Guo}]{dong2022protecting}
\bibinfo{author}{X.~Dong}, \bibinfo{author}{J.~Bao}, \bibinfo{author}{D.~Chen}, \bibinfo{author}{T.~Zhang}, \bibinfo{author}{W.~Zhang}, \bibinfo{author}{N.~Yu}, \bibinfo{author}{D.~Chen}, \bibinfo{author}{F.~Wen}, \bibinfo{author}{B.~Guo},
\newblock \bibinfo{title}{Protecting celebrities from deepfake with identity consistency transformer},
\newblock in: \bibinfo{booktitle}{Proceedings of the IEEE/CVF Conference on Computer Vision and Pattern Recognition}, \bibinfo{year}{2022}, pp. \bibinfo{pages}{9468--9478}.
\bibitem[{Bhaumik and Woo(2023)}]{bhaumik2023exploiting}
\bibinfo{author}{K.~K. Bhaumik}, \bibinfo{author}{S.~S. Woo},
\newblock \bibinfo{title}{Exploiting inconsistencies in object representations for deepfake video detection},
\newblock in: \bibinfo{booktitle}{Proceedings of the 2nd Workshop on Security Implications of Deepfakes and Cheapfakes}, \bibinfo{year}{2023}, pp. \bibinfo{pages}{11--15}.
\bibitem[{Zhao et~al.(2021)Zhao, Zhou, Chen, Wei, Zhang, and Yu}]{zhao2021multi}
\bibinfo{author}{H.~Zhao}, \bibinfo{author}{W.~Zhou}, \bibinfo{author}{D.~Chen}, \bibinfo{author}{T.~Wei}, \bibinfo{author}{W.~Zhang}, \bibinfo{author}{N.~Yu},
\newblock \bibinfo{title}{Multi-attentional deepfake detection},
\newblock in: \bibinfo{booktitle}{Proceedings of the IEEE/CVF conference on computer vision and pattern recognition}, \bibinfo{year}{2021}, pp. \bibinfo{pages}{2185--2194}.
\bibitem[{Tan et~al.(2023)Tan, Wang, Wang, Yang, Chen, and Guo}]{tan2023deepfake}
\bibinfo{author}{L.~Tan}, \bibinfo{author}{Y.~Wang}, \bibinfo{author}{J.~Wang}, \bibinfo{author}{L.~Yang}, \bibinfo{author}{X.~Chen}, \bibinfo{author}{Y.~Guo},
\newblock \bibinfo{title}{Deepfake video detection via facial action dependencies estimation},
\newblock in: \bibinfo{booktitle}{Proceedings of the AAAI Conference on Artificial Intelligence}, volume~\bibinfo{volume}{37}, \bibinfo{year}{2023}, pp. \bibinfo{pages}{5276--5284}.
\bibitem[{Dong et~al.(2023)Dong, Wang, Ji, Liang, Fan, and Ge}]{dong2023implicit}
\bibinfo{author}{S.~Dong}, \bibinfo{author}{J.~Wang}, \bibinfo{author}{R.~Ji}, \bibinfo{author}{J.~Liang}, \bibinfo{author}{H.~Fan}, \bibinfo{author}{Z.~Ge},
\newblock \bibinfo{title}{Implicit identity leakage: The stumbling block to improving deepfake detection generalization},
\newblock in: \bibinfo{booktitle}{Proceedings of the IEEE/CVF conference on computer vision and pattern recognition}, \bibinfo{year}{2023}, pp. \bibinfo{pages}{3994--4004}.
\bibitem[{Luo et~al.(2021)Luo, Zhang, Yan, and Liu}]{luo2021generalizing}
\bibinfo{author}{Y.~Luo}, \bibinfo{author}{Y.~Zhang}, \bibinfo{author}{J.~Yan}, \bibinfo{author}{W.~Liu},
\newblock \bibinfo{title}{Generalizing face forgery detection with high-frequency features},
\newblock in: \bibinfo{booktitle}{Proceedings of the IEEE/CVF conference on computer vision and pattern recognition}, \bibinfo{year}{2021}, pp. \bibinfo{pages}{16317--16326}.
\bibitem[{Gao et~al.(2024)Gao, Micheletto, Orr{\`u}, Concas, Feng, Marcialis, and Roli}]{gao2024texture}
\bibinfo{author}{J.~Gao}, \bibinfo{author}{M.~Micheletto}, \bibinfo{author}{G.~Orr{\`u}}, \bibinfo{author}{S.~Concas}, \bibinfo{author}{X.~Feng}, \bibinfo{author}{G.~L. Marcialis}, \bibinfo{author}{F.~Roli},
\newblock \bibinfo{title}{Texture and artifact decomposition for improving generalization in deep-learning-based deepfake detection},
\newblock \bibinfo{journal}{Engineering Applications of Artificial Intelligence} \bibinfo{volume}{133} (\bibinfo{year}{2024}) \bibinfo{pages}{108450}.
\bibitem[{Tan et~al.(2024)Tan, Zhao, Wei, Gu, Liu, and Wei}]{tan2024frequency}
\bibinfo{author}{C.~Tan}, \bibinfo{author}{Y.~Zhao}, \bibinfo{author}{S.~Wei}, \bibinfo{author}{G.~Gu}, \bibinfo{author}{P.~Liu}, \bibinfo{author}{Y.~Wei},
\newblock \bibinfo{title}{Frequency-aware deepfake detection: Improving generalizability through frequency space domain learning},
\newblock in: \bibinfo{booktitle}{Proceedings of the AAAI Conference on Artificial Intelligence}, volume~\bibinfo{volume}{38}, \bibinfo{year}{2024}, pp. \bibinfo{pages}{5052--5060}.
\bibitem[{Zhu et~al.(2024)Zhu, Zhang, Yin, Yin, and Lu}]{zhu2024deepfake}
\bibinfo{author}{C.~Zhu}, \bibinfo{author}{B.~Zhang}, \bibinfo{author}{Q.~Yin}, \bibinfo{author}{C.~Yin}, \bibinfo{author}{W.~Lu},
\newblock \bibinfo{title}{Deepfake detection via inter-frame inconsistency recomposition and enhancement},
\newblock \bibinfo{journal}{Pattern Recognition} \bibinfo{volume}{147} (\bibinfo{year}{2024}) \bibinfo{pages}{110077}.
\bibitem[{Zhao et~al.(2023)Zhao, Jin, Gao, Wu, Yao, and Jiang}]{zhao2023tan}
\bibinfo{author}{Y.~Zhao}, \bibinfo{author}{X.~Jin}, \bibinfo{author}{S.~Gao}, \bibinfo{author}{L.~Wu}, \bibinfo{author}{S.~Yao}, \bibinfo{author}{Q.~Jiang},
\newblock \bibinfo{title}{Tan-gfd: generalizing face forgery detection based on texture information and adaptive noise mining},
\newblock \bibinfo{journal}{Applied Intelligence}  (\bibinfo{year}{2023}) \bibinfo{pages}{1--21}.
\bibitem[{Yan et~al.(2024)Yan, Luo, Lyu, Liu, and Wu}]{yan2024transcending}
\bibinfo{author}{Z.~Yan}, \bibinfo{author}{Y.~Luo}, \bibinfo{author}{S.~Lyu}, \bibinfo{author}{Q.~Liu}, \bibinfo{author}{B.~Wu},
\newblock \bibinfo{title}{Transcending forgery specificity with latent space augmentation for generalizable deepfake detection},
\newblock in: \bibinfo{booktitle}{Proceedings of the IEEE/CVF Conference on Computer Vision and Pattern Recognition}, \bibinfo{year}{2024}, pp. \bibinfo{pages}{8984--8994}.
\bibitem[{Ding et~al.(2021)Ding, Zhang, Ma, Han, Ding, and Sun}]{ding2021repvgg}
\bibinfo{author}{X.~Ding}, \bibinfo{author}{X.~Zhang}, \bibinfo{author}{N.~Ma}, \bibinfo{author}{J.~Han}, \bibinfo{author}{G.~Ding}, \bibinfo{author}{J.~Sun},
\newblock \bibinfo{title}{Repvgg: Making vgg-style convnets great again},
\newblock in: \bibinfo{booktitle}{Proceedings of the IEEE/CVF conference on computer vision and pattern recognition}, \bibinfo{year}{2021}, pp. \bibinfo{pages}{13733--13742}.
\bibitem[{Wang et~al.(2004)Wang, Bovik, Sheikh, and Simoncelli}]{wang2004image}
\bibinfo{author}{Z.~Wang}, \bibinfo{author}{A.~C. Bovik}, \bibinfo{author}{H.~R. Sheikh}, \bibinfo{author}{E.~P. Simoncelli},
\newblock \bibinfo{title}{Image quality assessment: from error visibility to structural similarity},
\newblock \bibinfo{journal}{IEEE transactions on image processing} \bibinfo{volume}{13} (\bibinfo{year}{2004}) \bibinfo{pages}{600--612}.
\bibitem[{Loza et~al.(2006)Loza, Mihaylova, Canagarajah, and Bull}]{loza2006structural}
\bibinfo{author}{A.~Loza}, \bibinfo{author}{L.~Mihaylova}, \bibinfo{author}{N.~Canagarajah}, \bibinfo{author}{D.~Bull},
\newblock \bibinfo{title}{Structural similarity-based object tracking in video sequences},
\newblock in: \bibinfo{booktitle}{2006 9th International Conference on Information Fusion}, \bibinfo{organization}{IEEE}, \bibinfo{year}{2006}, pp. \bibinfo{pages}{1--6}.
\bibitem[{Hou et~al.(2020)Hou, Zhang, Cheng, and Feng}]{hou2020strip}
\bibinfo{author}{Q.~Hou}, \bibinfo{author}{L.~Zhang}, \bibinfo{author}{M.-M. Cheng}, \bibinfo{author}{J.~Feng},
\newblock \bibinfo{title}{Strip pooling: Rethinking spatial pooling for scene parsing},
\newblock in: \bibinfo{booktitle}{Proceedings of the IEEE/CVF conference on computer vision and pattern recognition}, \bibinfo{year}{2020}, pp. \bibinfo{pages}{4003--4012}.
\bibitem[{Ahmed et~al.(1974)Ahmed, Natarajan, and Rao}]{ahmed1974discrete}
\bibinfo{author}{N.~Ahmed}, \bibinfo{author}{T.~Natarajan}, \bibinfo{author}{K.~R. Rao},
\newblock \bibinfo{title}{Discrete cosine transform},
\newblock \bibinfo{journal}{IEEE transactions on Computers} \bibinfo{volume}{100} (\bibinfo{year}{1974}) \bibinfo{pages}{90--93}.
\bibitem[{Rao and Yip(2014)}]{rao2014discrete}
\bibinfo{author}{K.~R. Rao}, \bibinfo{author}{P.~Yip}, \bibinfo{title}{Discrete cosine transform: algorithms, advantages, applications}, \bibinfo{publisher}{Academic press}, \bibinfo{year}{2014}.
\bibitem[{Braginsky and Khalili(1995)}]{braginsky1995quantum}
\bibinfo{author}{V.~B. Braginsky}, \bibinfo{author}{F.~Y. Khalili}, \bibinfo{title}{Quantum measurement}, \bibinfo{publisher}{Cambridge University Press}, \bibinfo{year}{1995}.
\bibitem[{Jacobs and Steck(2006)}]{jacobs2006straightforward}
\bibinfo{author}{K.~Jacobs}, \bibinfo{author}{D.~A. Steck},
\newblock \bibinfo{title}{A straightforward introduction to continuous quantum measurement},
\newblock \bibinfo{journal}{Contemporary Physics} \bibinfo{volume}{47} (\bibinfo{year}{2006}) \bibinfo{pages}{279--303}.
\bibitem[{Rossler et~al.(2019)Rossler, Cozzolino, Verdoliva, Riess, Thies, and Nie{\ss}ner}]{rossler2019faceforensics++}
\bibinfo{author}{A.~Rossler}, \bibinfo{author}{D.~Cozzolino}, \bibinfo{author}{L.~Verdoliva}, \bibinfo{author}{C.~Riess}, \bibinfo{author}{J.~Thies}, \bibinfo{author}{M.~Nie{\ss}ner},
\newblock \bibinfo{title}{Faceforensics++: Learning to detect manipulated facial images},
\newblock in: \bibinfo{booktitle}{Proceedings of the IEEE/CVF international conference on computer vision}, \bibinfo{year}{2019}, pp. \bibinfo{pages}{1--11}.
\bibitem[{Li et~al.(2020)Li, Yang, Sun, Qi, and Lyu}]{li2020celeb}
\bibinfo{author}{Y.~Li}, \bibinfo{author}{X.~Yang}, \bibinfo{author}{P.~Sun}, \bibinfo{author}{H.~Qi}, \bibinfo{author}{S.~Lyu},
\newblock \bibinfo{title}{Celeb-df: A large-scale challenging dataset for deepfake forensics},
\newblock in: \bibinfo{booktitle}{Proceedings of the IEEE/CVF conference on computer vision and pattern recognition}, \bibinfo{year}{2020}, pp. \bibinfo{pages}{3207--3216}.
\bibitem[{Dolhansky et~al.(2020)Dolhansky, Bitton, Pflaum, Lu, Howes, Wang, and Ferrer}]{dolhansky2020deepfake}
\bibinfo{author}{B.~Dolhansky}, \bibinfo{author}{J.~Bitton}, \bibinfo{author}{B.~Pflaum}, \bibinfo{author}{J.~Lu}, \bibinfo{author}{R.~Howes}, \bibinfo{author}{M.~Wang}, \bibinfo{author}{C.~C. Ferrer},
\newblock \bibinfo{title}{The deepfake detection challenge (dfdc) dataset},
\newblock \bibinfo{journal}{arXiv preprint arXiv:2006.07397}  (\bibinfo{year}{2020}).
\bibitem[{Jiang et~al.(2020)Jiang, Li, Wu, Qian, and Loy}]{jiang2020deeperforensics}
\bibinfo{author}{L.~Jiang}, \bibinfo{author}{R.~Li}, \bibinfo{author}{W.~Wu}, \bibinfo{author}{C.~Qian}, \bibinfo{author}{C.~C. Loy},
\newblock \bibinfo{title}{Deeperforensics-1.0: A large-scale dataset for real-world face forgery detection},
\newblock in: \bibinfo{booktitle}{Proceedings of the IEEE/CVF conference on computer vision and pattern recognition}, \bibinfo{year}{2020}, pp. \bibinfo{pages}{2889--2898}.
\bibitem[{Nadimpalli and Rattani(2022)}]{nadimpalli2022improving}
\bibinfo{author}{A.~V. Nadimpalli}, \bibinfo{author}{A.~Rattani},
\newblock \bibinfo{title}{On improving cross-dataset generalization of deepfake detectors},
\newblock in: \bibinfo{booktitle}{Proceedings of the IEEE/CVF conference on computer vision and pattern recognition}, \bibinfo{year}{2022}, pp. \bibinfo{pages}{91--99}.
\bibitem[{Deng et~al.(2009)Deng, Dong, Socher, Li, Li, and Fei-Fei}]{deng2009imagenet}
\bibinfo{author}{J.~Deng}, \bibinfo{author}{W.~Dong}, \bibinfo{author}{R.~Socher}, \bibinfo{author}{L.-J. Li}, \bibinfo{author}{K.~Li}, \bibinfo{author}{L.~Fei-Fei},
\newblock \bibinfo{title}{Imagenet: A large-scale hierarchical image database},
\newblock in: \bibinfo{booktitle}{2009 IEEE/CVF conference on computer vision and pattern recognition}, \bibinfo{organization}{Ieee}, \bibinfo{year}{2009}, pp. \bibinfo{pages}{248--255}.
\bibitem[{Kingma and Ba(2014)}]{kingma2014adam}
\bibinfo{author}{D.~P. Kingma}, \bibinfo{author}{J.~Ba},
\newblock \bibinfo{title}{Adam: A method for stochastic optimization},
\newblock \bibinfo{journal}{arXiv preprint arXiv:1412.6980}  (\bibinfo{year}{2014}).
\bibitem[{Sun et~al.(2022)Sun, Yao, Chen, Ding, Li, and Ji}]{sun2022dual}
\bibinfo{author}{K.~Sun}, \bibinfo{author}{T.~Yao}, \bibinfo{author}{S.~Chen}, \bibinfo{author}{S.~Ding}, \bibinfo{author}{J.~Li}, \bibinfo{author}{R.~Ji},
\newblock \bibinfo{title}{Dual contrastive learning for general face forgery detection},
\newblock in: \bibinfo{booktitle}{Proceedings of the AAAI conference on artificial intelligence}, volume~\bibinfo{volume}{36}, \bibinfo{year}{2022}, pp. \bibinfo{pages}{2316--2324}.
\bibitem[{Selvaraju et~al.(2017)Selvaraju, Cogswell, Das, Vedantam, Parikh, and Batra}]{selvaraju2017grad}
\bibinfo{author}{R.~R. Selvaraju}, \bibinfo{author}{M.~Cogswell}, \bibinfo{author}{A.~Das}, \bibinfo{author}{R.~Vedantam}, \bibinfo{author}{D.~Parikh}, \bibinfo{author}{D.~Batra},
\newblock \bibinfo{title}{Grad-cam: Visual explanations from deep networks via gradient-based localization},
\newblock in: \bibinfo{booktitle}{Proceedings of the IEEE international conference on computer vision}, \bibinfo{year}{2017}, pp. \bibinfo{pages}{618--626}.
\bibitem[{Van~der Maaten and Hinton(2008)}]{van2008visualizing}
\bibinfo{author}{L.~Van~der Maaten}, \bibinfo{author}{G.~Hinton},
\newblock \bibinfo{title}{Visualizing data using t-sne.},
\newblock \bibinfo{journal}{Journal of machine learning research} \bibinfo{volume}{9} (\bibinfo{year}{2008}).
\bibitem[{Koonce and Koonce(2021)}]{koonce2021efficientnet}
\bibinfo{author}{B.~Koonce}, \bibinfo{author}{B.~Koonce},
\newblock \bibinfo{title}{Efficientnet},
\newblock \bibinfo{journal}{Convolutional neural networks with swift for Tensorflow: image recognition and dataset categorization}  (\bibinfo{year}{2021}) \bibinfo{pages}{109--123}.
\bibitem[{Duan et~al.(2023)Duan, Long, Wang, Zhang, Willcocks, and Shao}]{duan2023dynamic}
\bibinfo{author}{H.~Duan}, \bibinfo{author}{Y.~Long}, \bibinfo{author}{S.~Wang}, \bibinfo{author}{H.~Zhang}, \bibinfo{author}{C.~G. Willcocks}, \bibinfo{author}{L.~Shao},
\newblock \bibinfo{title}{Dynamic unary convolution in transformers},
\newblock \bibinfo{journal}{IEEE Transactions on Pattern Analysis and Machine Intelligence} \bibinfo{volume}{45} (\bibinfo{year}{2023}) \bibinfo{pages}{12747--12759}.
\bibitem[{Miao et~al.(2025)Miao, Duan, Bai, Shah, Song, Long, Ranjan, and Shao}]{miao2025laser}
\bibinfo{author}{X.~Miao}, \bibinfo{author}{H.~Duan}, \bibinfo{author}{Y.~Bai}, \bibinfo{author}{T.~Shah}, \bibinfo{author}{J.~Song}, \bibinfo{author}{Y.~Long}, \bibinfo{author}{R.~Ranjan}, \bibinfo{author}{L.~Shao},
\newblock \bibinfo{title}{Laser: Efficient language-guided segmentation in neural radiance fields},
\newblock \bibinfo{journal}{arXiv preprint arXiv:2501.19084}  (\bibinfo{year}{2025}).

\end{thebibliography}


\end{document}